\documentclass{article} 

\usepackage{arxiv}
\usepackage{amsmath,amssymb,amsfonts}
\usepackage{algorithmic}
\usepackage{multirow}
\usepackage{booktabs}

\usepackage{graphicx}
\graphicspath{{./pdf/}{./jpeg/}{./png/}}
\usepackage[T1]{fontenc}
\usepackage[utf8]{inputenc}
\usepackage{nicefrac}
\usepackage{hyperref}
\usepackage{array}
\DeclareGraphicsExtensions{.pdf,.jpeg,.png}

\usepackage{textcomp}
\usepackage{xcolor}
\usepackage{verbatim}
\usepackage{algorithm}
\usepackage{rotating}
\usepackage{float}
\usepackage{stackrel}
\usepackage{subscript}
\usepackage[caption=false]{subfig}
\usepackage{url}
\usepackage{comment}

\usepackage{amsthm}
\theoremstyle{definition}
\newtheorem{definition}{Definition}[section]

\AtBeginDocument{%
\abovedisplayskip=-30pt
\abovedisplayshortskip=-10pt
}

\begin{document}

\title{Overlap Number of Balls Model-Agnostic CounterFactuals (ONB-MACF): A Data-Morphology-based Counterfactual Generation Method for Trustworthy Artificial Intelligence}

\author{Jos{\'e} Daniel Pascual-Triana\thanks{Andalusian Institute of Data Science and Computational Intelligence (DASCI), University of Granada, Granada, 18071, Spain.}\> \thanks{Corresponding authors: jdpascualt@ugr.es, alfh@ugr.es.} \\ \and \textbf{Alberto Fern{\'a}ndez}\label{cor2}\footnotemark[1]\> \footnotemark[2] \\ \and \textbf{Javier Del Ser} \footnotemark[1]\> \thanks{TECNALIA, Basque Research \& Technology Alliance (BRTA), Derio, Spain.}\> \thanks{University of the Basque Country (UPV/EHU), Bilbao, Spain.} \\ \and \textbf{Francisco Herrera}\footnotemark[1]}

\date{Received: date / Accepted: date}
\maketitle

\begin{abstract}

Explainable Artificial Intelligence (XAI) is a pivotal research domain aimed at understanding the operational mechanisms of AI systems, particularly those considered “black boxes” due to their complex, opaque nature. XAI seeks to make these AI systems more understandable and trustworthy, providing insight into their decision-making processes. By producing clear and comprehensible explanations, XAI enables users, practitioners, and stakeholders to trust a model’s decisions.  This work analyses the value of data morphology strategies in generating counterfactual explanations.  It introduces the Overlap Number of Balls Model-Agnostic CounterFactuals (ONB-MACF) method, a model-agnostic counterfactual generator that leverages data morphology to estimate a model’s decision boundaries. The ONB-MACF method constructs hyperspheres in the data space whose covered points share a class, mapping the decision boundary. Counterfactuals are then generated by incrementally adjusting an instance’s attributes towards the nearest alternate-class hypersphere, crossing the decision boundary with minimal modifications. By design, the ONB-MACF method generates feasible and sparse counterfactuals that follow the data distribution. Our comprehensive benchmark from a double perspective (quantitative and qualitative) shows that the ONB-MACF method outperforms existing state-of-the-art counterfactual generation methods across multiple quality metrics on diverse tabular datasets. This supports our hypothesis, showcasing the potential of data-morphology-based explainability strategies for trustworthy AI.

\keywords{Explainable Artificial Intelligence \and Model-agnostic Explanations \and Counterfactual Analysis \and Data Morphology \and Trustworthy Artificial Intelligence.}

\end{abstract}

\section{Introduction}

The expansive application of AI systems underscores the necessity for these systems to not only perform with exceptional accuracy but to do so with transparency, auditability, and accountability at every stage of their life-cycle \cite{Diaz-Rodriguez2023} —from data collection and model training to deployment and ongoing validation. Regulatory frameworks like the European Union's Artificial Intelligence Act\footnote{AI Act, \url{https://artificialintelligenceact.eu/}, last accessed 25 April 2024} \cite{Veale21} and initiatives such as the United States National Institute of Standards and Technology's (NIST) framework for AI\footnote{NIST Framework for AI \url{https://www.nist.gov/artificial-intelligence}, last accessed 12 April 2024}, emphasise the critical need for 
AI technologies that are innovative, explainable, responsible and reliable.

In this context, Explainable AI (XAI) is an essential research area whose purpose is to shed light on the inner workings of complex AI systems to a given audience \cite{barredo_arrieta_explainable_2020, LONGO2024102301}. With the clarification of AI decision-making processes, XAI seeks to improve their transparency and understandability \cite{Ding2022238}.
XAI encompasses a broad spectrum of methodologies, ranging from inherently transparent models that allow for straightforward interpretation of their processes, to post-hoc explanatory tools designed to elucidate the workings of otherwise opaque models \cite{Minh20223503, Ali2023}. Moreover, the field distinguishes between model-specific 
 explainers, tailored for specific types of models \cite{karimi_model-agnostic_2020}, but also model-agnostic ones, which provide explanations across different types of classification models.


To elucidate the decision-making processes, post-hoc explanations through factual, semifactual, and counterfactual instances stand out for their 
ability to validate models \cite{ribeiro_why_2016}. Factual explanations delineate reasons behind a model's decision for a specific case 
\cite{fernandez_factual_2022}. Semifactual explanations explore hypothetical scenarios where a sample's class remains unchanged despite feasible modifications, offering 
understanding of the model's response to input changes \cite{aryal_even_2023, r_fernandez_explanation_2022}. 
Counterfactual explanations pivot on the minimal feasible changes needed to alter a model's decision, providing 
dual utility: they empower users seeking to understand or challenge decisions and facilitate model auditing by revealing feature importance and potential biases or unrealistic feature values \cite{guidotti_counterfactual_2022}.

Counterfactual explanations hinge on a suite of features that collectively determine their quality and utility \cite{DelSer2024}. Primarily, a counterfactual must belong to a different class, providing an alternative to the model's decision. Minimalistic intervention advocates for altering a concise set of features to avoid unnecessary complexity and enhance interpretability. The counterfactual must be close to the original instance (involving minimal yet effective changes), realistic and representative of the data distribution. It should only alter features that can be realistically modified, avoiding immutable characteristics. Moreover, offering multiple pathways for action or understanding in generated counterfactuals allows users to evaluate their preferred course of action.

Despite their utility, current approaches to generating counterfactuals face  
limitations, as discussed in \cite{stepin_survey_2021}. Many rely on model-specific methodologies, restricting their applicability 
\cite{chou_counterfactuals_2022}. Even when the design is 
model-agnostic, it does not always properly estimate the class boundaries \cite{keane_if_2021}, which complicates the generation of close, minimal-change counterfactuals
that adhere to actual data distributions. Furthermore, existing techniques often overlook the treatment of protected or immutable features, rendering some counterfactuals impractical or unethical \cite{Kasirzadeh2021228}
. 
Finally, the inability to provide sets of diverse counterfactuals when needed restricts the depth of insight into the model's behaviour across possible outcomes.


Thus, our hypothesis is based on the potential usefulness of data-morphology-based explainability strategies 
for trustworthy AI and, in particular,  in the generation of counterfactual explanations. This hypothesis is associated with a method that embodies the ONB methodology \cite{pascual-triana_revisiting_2021}, derived from geometry-based mathematical tools \cite{manukyan_classification_2016} and the morphology-based complexity metric, which facilitates coverings of tabular datasets by encapsulating data clusters into ``balls'' (the interior of hyper-spheres).

This work proposes the Overlap Number of Balls Model Agnostic CounterFactual method (ONB-MACF), a novel explanation strategy leveraging data morphology to delineate a model's class boundaries\footnote{The code concerning the ONB-MACF method can be provided upon request.}. It is based on the construction of the aforementioned balls and involves a mapping strategy, where each ball encompasses instances from a single class based on the classifier's predictions and a predefined distance function. This mapping not only approximates complex class boundaries but also identifies pathways for counterfactual transitions, leveraging the proximity and geometry of balls associated with alternative class labels. When a counterfactual is sought for a specific input data instance, our approach harnesses this geometrically enriched map to navigate towards the most plausible counterfactual candidates, starting from the boundary closest to the original instance and moving towards the projected centre of an adjacent class's ball.

By construction, ONB-MACF is a model-agnostic counterfactual explainer adaptable to class boundary irregularities. Utilizing the ball boundaries as a basis, the methodology can simultaneously maximise different requirements. Specifically, we ensure the feasibility and foster 
minimal feature modification and low distance from the original sample. This is achieved when carrying out small modifications towards the projected ball centre, obtained after enforcing immutable feature constraints, 
while mitigating unnecessary feature alterations
. Finally, the methodology's prowess extends to generating plausible semifactuals, offering an additional layer of explainability.


To analyse the good capabilities of the ONB-MACF method, we conduct a comprehensive evaluation from a double perspective, both quantitatively and qualitatively. Our study employs pre-trained neural networks as baseline classification models for 8 classification problems, each with distinct characteristics and widely recognised in counterfactual studies. Comparison methodologies include a wide selection of explainers renowned for their efficacy in counterfactual generation \cite{pawelczyk_carla_2021}. Amongst them, we highlight the use of Growing Spheres \cite{laugel_comparison-based_2018} and NICE \cite{brughmans_nice_2023}, methods that share similarities with the ONB-MACF method in their reliance on hyperspheres and prototypes, respectively. Despite their points in common, ONB-MACF makes use of hyperspheres' morphology, instead of employing them as a limit for the random candidate generation of Growing Spheres, and creates its own (closer) prototypes in addition to existing ones, while NICE only uses the latter. Quantitative performance is measured across a set of well-established metrics \cite{verma_counterfactual_2020}, each tailored to assess critical aspects of counterfactual quality; the results indicate that counterfactuals generated using the ONB-MACF method are close and in distribution, involve few feature changes and avoid changes in immutable features. Likewise, a complementary qualitative analysis is carried out to show the good behaviour of our approach and the usefulness of the provided explanations, such as using the appropriate features for the class changes and avoiding the protected features.

The rest of this paper is structured in the following manner. Firstly, the preliminaries and state of the art on counterfactual explanations will be introduced in Section \ref{prelim}. Then, the proposed ONB-MACF method will be detailed in Section \ref{ONB-MACF}. Next, the experimental framework for the comparison against the state of the art will be described in Section \ref{frame}. Afterwards, the quantitative analysis from the comparison of ONB-MACF with other benchmark methods will be explained in Section \ref{exper} (per-dataset results tables are shown in \ref{result_tables}). The qualitative analysis of the ONB-MACF method and its comparison to the aforementioned methods will be provided in Section \ref{qualitative}. Finally, the concluding remarks and some paths for future work will be presented in Section \ref{concl}.

\section{Preliminaries on Counterfactual Explanations}\label{prelim}

The structure of this section, which serves as an introduction to counterfactual explanations, is the following. Firstly, the basics of counterfactual explanations, their concept, benefits and expected qualities, are explained in Section \ref{basics}; then, a reference to the related works is included in Section \ref{subsec:related}.

\subsection{Fundamentals of Counterfactual Explanations}\label{basics}

Counterfactual explanations are a technique to explain decisions whose basis is a generated example with changes in some features compared to the initial datum to explain, such that the classes of both instances differ. A more formal definition of counterfactual explanations is given in Definition \ref{def_count}.

\begin{definition}\label{def_count}
Given a sample $x \in X$ from data space $X$ and a classification model $f: X \xrightarrow{} Y$,  where $Y$ is the class set and $|Y|=n \in \mathbb{N}$ and $f(x)=y \in Y$, a counterfactual of $x$ would be $x' \in X$ where $|x-x'|<\delta$, where $\delta$ is small and $f(x')=y' \in Y, y\neq y'$. 
\end{definition}

Counterfactuals began to take importance for machine learning as an explanation technique due to, amongst other characteristics, their aforementioned aptitudes for challenging decisions and closeness to human thinking.

In fact, this type of explanation is usually considered understandable by humans, as it is contrastive (it gives information about the critical features for the decision) and can be expressed in a few terms involving data features and values, which, most often, use natural language. Moreover, not only does it give insight into how the decision was made, but it also indicates which kinds of small changes would lead to a different outcome in an intuitive way. Thus, users know what they should modify in order to challenge the decision that, for example, denies them a loan. Furthermore, they can be a precious tool in auditing models, as a developer can easily read the explanation and see whether the model uses the right features to decide the class change. This is particularly important from the fairness and ethics standpoints, as the presence of sensitive/protected features in a counterfactual explanation can explicitly point out classification bias, which could lead to a need for accountability\footnote{AI Act, \url{https://artificialintelligenceact.eu/}, last accessed 25 April 2024}. 
For the effectiveness of counterfactuals as high-quality, understandable explanation techniques, there are several recommended or even mandatory characteristics to fulfill
\cite{guidotti_counterfactual_2022}, \cite{verma_counterfactual_2020}, \cite{chou_counterfactuals_2022}, from which the objective ones, and therefore those that can mostly be  measured, are enumerated below.

\begin{itemize}
    \item Validity: the counterfactual's class must differ from the sample's.  
    \item Sparsity: a good counterfactual should change few features. There is no definitive consensus on the optimum number of changes, but it is recommended to change only 1-3 features
    \cite{keane_if_2021}.
    \item Similarity: a counterfactual must be close to the sample, and this closeness could be measured using different distance metrics.
    \item Actionability: a good counterfactual must not change features that cannot be acted upon, such as immutable features.    
    \item Plausibility: a counterfactual should be in-distribution according to the observable data and it should not be considered an outlier.
    \item Causality: a good counterfactual should respect the causal relations from the dataset, as otherwise, it might not be plausible.     
    \item Diversity: if multiple counterfactuals are given for one sample, they should be as diverse as possible,  changing different features so that distinct actions can be taken.
\end{itemize}

\subsection{Related Work}\label{subsec:related}

This section presents a brief reference to the state of the art on counterfactual explanations. The counterfactual methods are organised with respect to different key aspects such as model specificity, technique, information requirements (from data or model) and target data type. Should the reader want to know more about individual topics, there have been multiple surveys on the matter in recent years \cite{stepin_survey_2021,guidotti_counterfactual_2022,verma_counterfactual_2020,chou_counterfactuals_2022,keane_if_2021}.

According to model specificity, some techniques require a particular type (such as a decision tree or an ensemble \cite{karimi_model-agnostic_2020}) due to using the model's structure and parameters. Others accept any differentiable model, like  Diverse Counterfactual Explanations (DiCE) \cite{mothilal_explaining_2020}, which applies optimisation techniques based on cost functions to generate sets of feasible and diverse feature changes to swap an instance's class.  
Other methods work on any model (model-agnostic approaches), like  Feasible and Actionable Counterfactual Explanations (FACE) \cite{poyiadzi_face_2020}, which finds viable paths connecting instances from different classes in graphs generated using kNN or $\epsilon$-graphs. 
Model-specific approaches harness the intrinsic characteristics of the model to their advantage in exchange for being useful in fewer situations. Lately, model-agnostic approaches have been favoured due to their universality \cite{guidotti_counterfactual_2022}. 

Regarding the techniques for their implementation, several strategies have been used. Most strategies are based on optimisation techniques. Some of them are cost-based, like Contrastive Explanations Method (CEM) \cite{dhurandhar_explanations_2018}, which uses iterative algorithms to detect which attribute values are positively and negatively related to a class. Other optimisation techniques are restriction-based, like the aforementioned DiCE \cite{mothilal_explaining_2020}, or probability-based, like Counterfactual Latent Uncertainty Explanations (CLUE) \cite{antoran_getting_2021}, which employs probabilistic differentiable models to study how instance changes (respecting data distributions) increase the probability of it belonging to a different class. There are also heuristic approaches like Growing Spheres (GS) \cite{laugel_comparison-based_2018}, which generates uniformly distributed new instances on growing hyperspheres centred on the given instance. Less-explored strategies are instance-based. These can involve prototypes, like NICE \cite{brughmans_nice_2023}, whose strategy is to swap attribute values between the instance and members of the opposite class until the class changes. Other instance-based strategies use distances or graphs \cite{poyiadzi_face_2020}) or involve surrogate models, like in \cite{na_toward_2023}, where they approximate the model boundaries by learning a generative adversarial network and generate counterfactuals linearly in the boundary's direction.

Distinct counterfactual explainers also need access to different information. Some models might require access to training data (for example, instance-based explainers \cite{brughmans_nice_2023}). While all counterfactual explainers need access to parts of the prediction model, some of them require all information within the model \cite{karimi_model-agnostic_2020}, whereas others, 
such as the one proposed in \cite{wachter_counterfactual_2017}, use optimisation techniques with weights and the restrictions and, thus, need access to the gradients. Some model-agnostic methods only require the outputs from the prediction function, like Counterfactual Recourse Using Disentangled Subspaces (CRUDS) \cite{downs_cruds_nodate}, which uses autoencoders to obtain feasible and interpretable feature changes for an instance that modify its class while respecting causality and other restrictions. 

Besides that, not all counterfactuals are created with the same target data. 
Counterfactual generators have generally been designed to explain models that work on tabular data \cite{brughmans_nice_2023}. 
Some methods explain images, such as in \cite{DelSer2024}, where a generative adversarial network is used to produce counterfactuals with a focus on plausibility, the intensity of changes and adversarial power, or in \cite{na_toward_2023}, where the disentangled latent space of a generative adversarial network is employed to modify the instance towards the other class linearly. In a less explored manner, some methods have been designed for texts, such as in \cite{wu_polyjuice_2021}, where sentences can be modified in diverse ways using GPT-2, or in \cite{mollas_lionets_2023}, where neural network characteristics and feature importance are used to generate counterfactual words in sentences. While some explainers work for any of them \cite{poyiadzi_face_2020, de_oliveira_model-agnostic_2023}, most have been created for specific types to preserve their simplicity.

Since the quality of counterfactual explanations can be gauged in terms of many different metrics, it is very hard for a counterfactual generator to be considered superior to any other in all aspects simultaneously. While DiCE \cite{mothilal_explaining_2020} can obtain multiple counterfactuals that are different enough from each other, sometimes the boundary directions it chooses might not be optimal for single explanations. FACE \cite{poyiadzi_face_2020} uses graphs to find feasible changes, but it happens to the detriment of counterfactual distance. CEM \cite{dhurandhar_explanations_2018} does not include restrictions regarding which features can be changed, giving rise to unfeasible counterfactuals. Despite considering them, CLUE \cite{antoran_getting_2021} does not correctly enforce actionability constraints. The randomness of Growing spheres affects the method's robustness and can lead to counterfactuals that change all actionable attributes, which is unfeasible for the user. While  NICE \cite{brughmans_nice_2023} is fast, due to simply swapping attribute values between instances, it could get lower distances by employing the class boundary between both points. Wachter \cite{wachter_counterfactual_2017} optimises the distance without limiting the number of attributes to change, which can lead to unfeasible counterfactuals. CRUDS \cite{downs_cruds_nodate} disregards the geometry of the class boundaries, which can lead to counterfactuals further from the sample.

\section{Overlap Number of Balls Model-Agnostic CounterFactuals} \label{ONB-MACF}

This section presents ``Overlap Number of Balls Model Agnostic CounterFactual'' (ONB-MACF), a novel, model-agnostic, prototype-based counterfactual method developed for usage on tabular data. Unlike other counterfactual methods, ONB-MACF employs data morphology, which helps delineate the class boundaries amongst the classes. This fact helps the ONB-MACF method generate close, sparse counterfactuals that follow the data distribution.

Firstly, the algorithm's inner workings are presented in Section \ref{structure}; then, the advantages of the process followed are stated in Section \ref{advantages}.

\subsection{Structure of the ONB-MACF method}\label{structure}

The ONB-MACF method generates counterfactual explanations using a data-covering strategy from the Pure Class Cover Catch Digraph (PCCC-D) and ONB algorithms ~\cite{manukyan_classification_2016, pascual-triana_revisiting_2021}, so it requires access to the classification dataset, a distance function and the classifier's prediction function. For categorical features, the distance calculation involves using their one-hot encodings unless they are ordinal.

\begin{figure}[!ht]
\centering
\includegraphics[width=0.8\linewidth]{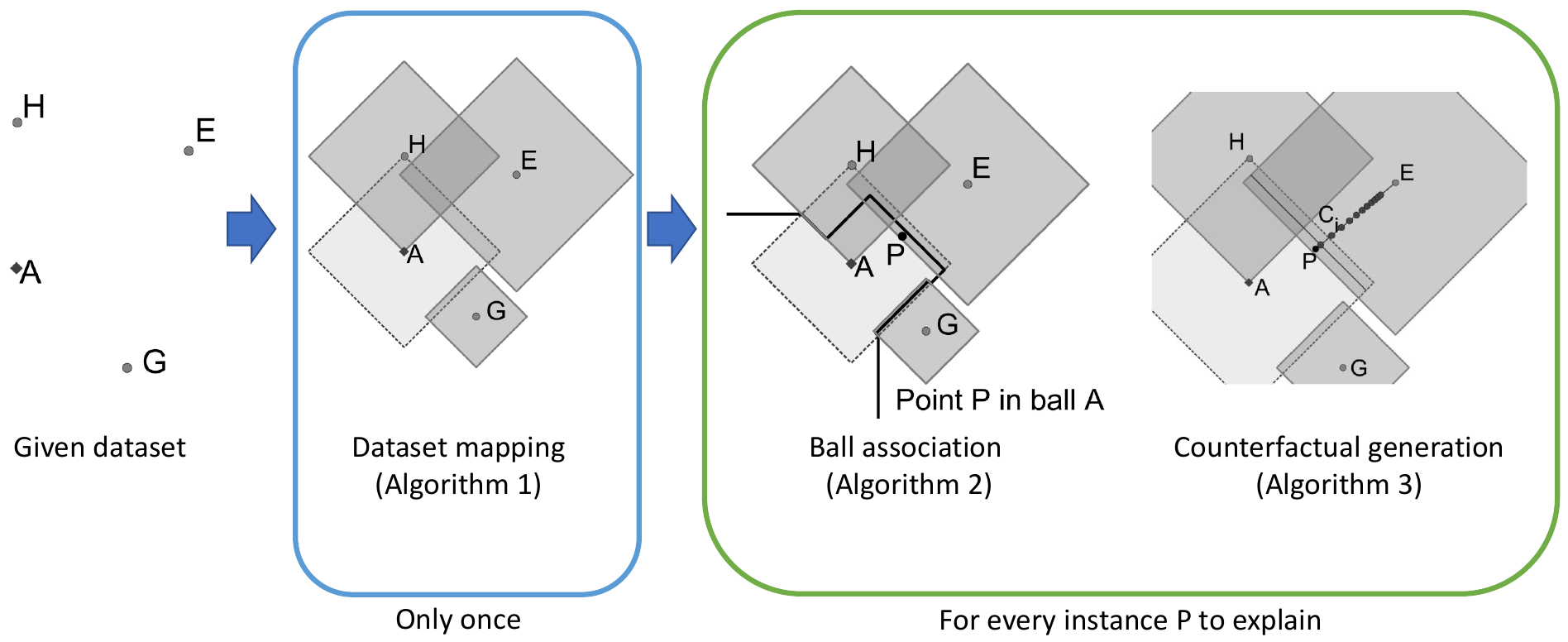}
\caption{\label{fig:diagrama}Given a dataset (represented by points $A$ to $H$), a class coverage mapping is obtained for a given distance metric (i.e. Manhattan); then, for each instance whose counterfactual is requested ($P$), said instance is assigned to an appropriate ball from the coverage ($A$), which helps select the closest class boundary ($A-E$), and counterfactual candidates ($C_{i}$) are generated between the boundary and the projected centre of the chosen ball of opposing class.}
\end{figure}

This strategy generates a ball coverage of the data where only elements from one class are present in each ball. 
It has proven useful when evaluating the complexity of datasets and their degrees of class overlap. This characterisation of the areas with class overlap (borderline areas) allows for the mapping of the approximate class boundaries, which might otherwise be very complex depending on the classification model.

Once a counterfactual is required for a given instance, the mapping helps to detect the directions in which better paths lie towards different classes. For that purpose, the ball boundaries and the centres of balls that are close to the instance but from the target classes are harnessed to generate close and feasible counterfactual candidates.

The ONB-MACF method can be structured in 3 stages:
\begin{enumerate}
    \item dataset mapping via the ONB algorithm for the characterisation of the classification data space (Algorithm \ref{alg1});
    \item instance association to the closest ball from the mapping for the selection of relevant boundaries for each instance (Algorithm \ref{alg2}) and
    \item counterfactual candidate generation for the obtaining of the chosen counterfactuals (Algorithm \ref{alg3}).
\end{enumerate}

To better understand this structure, a diagram is presented in Figure \ref{fig:diagrama}.

In Sections \ref{subsubsec:alg1} to \ref{subsubsec:alg3}, the three stages will be described, along with their associated algorithms.

\subsubsection{Dataset mapping via the ONB algorithm (Algorithm \ref{alg1})}\label{subsubsec:alg1}

In this stage, the ONB algorithm is used to cover all samples of the dataset using class-dependent open balls centred on instances from the dataset.

The coverage strategy is geometrically intuitive. For each class, the instances of that class and opposing ones are identified (lines 4 to 5); then, until all points of that class are covered, for each instance of that class, the ball centred on the instance with the maximum radius excluding instances of a different class is generated (line 9), the still uncovered instances that ball would cover (if it were to be chosen for the coverage) are obtained (line 10) and, if it would cover more instances than the previous temporarily selected ball, it becomes temporarily selected (lines 11 to 14); after all instances of that class have been checked, the ball that included the most instances not yet covered by balls in the coverage set (the chosen balls for the mapping) is selected and included in the coverage set (line 16), the set of points of that class that are still uncovered is updated (line 17) and the process continues until all instances of all classes are covered.

\begin{figure}[!ht]
\centering

    \subfloat[]{\label{ejbolasa}
    \includegraphics[width=.25\linewidth]{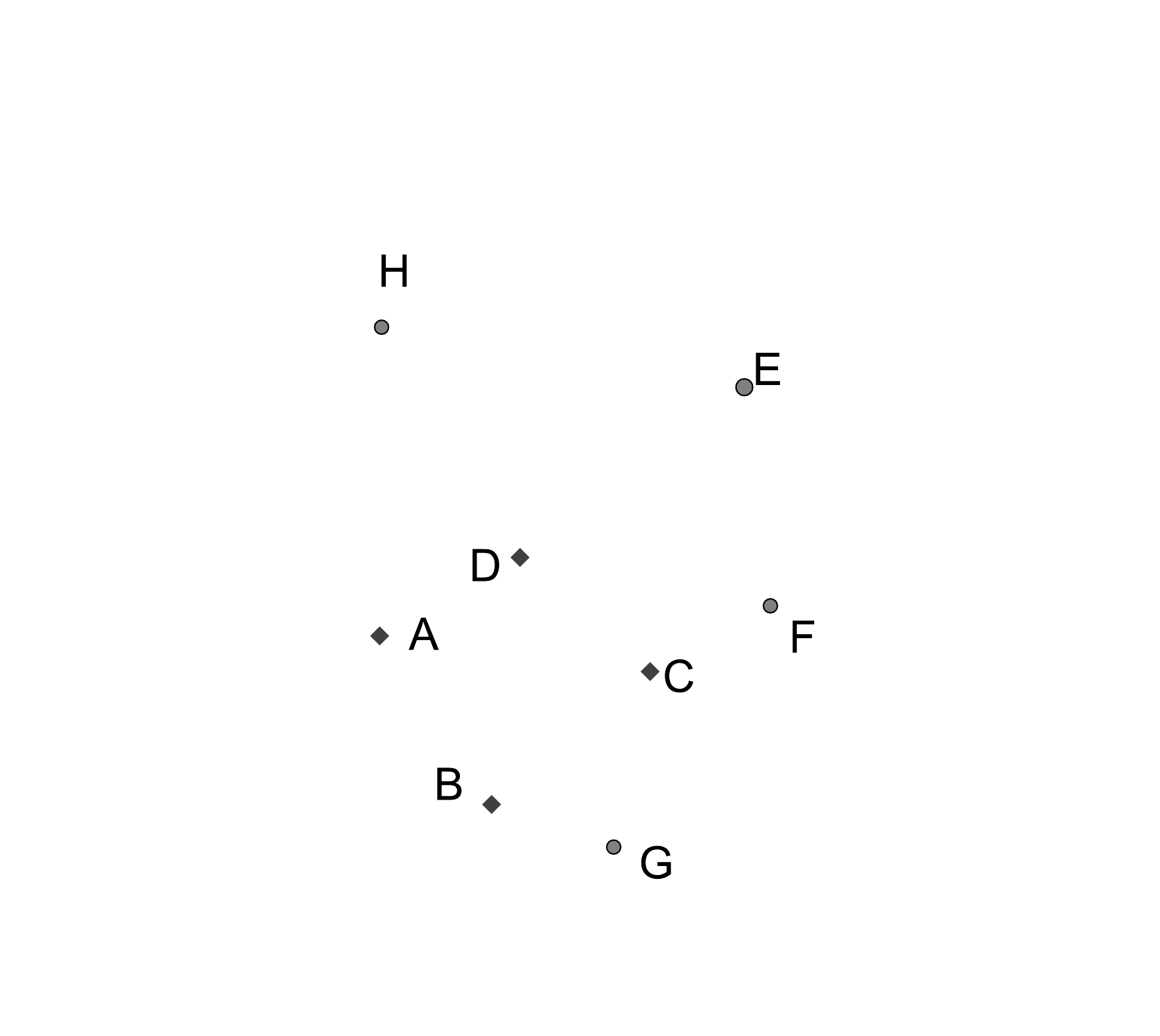}}
    \subfloat[]{\label{ejbolasb}
    \includegraphics[width=.25\linewidth]{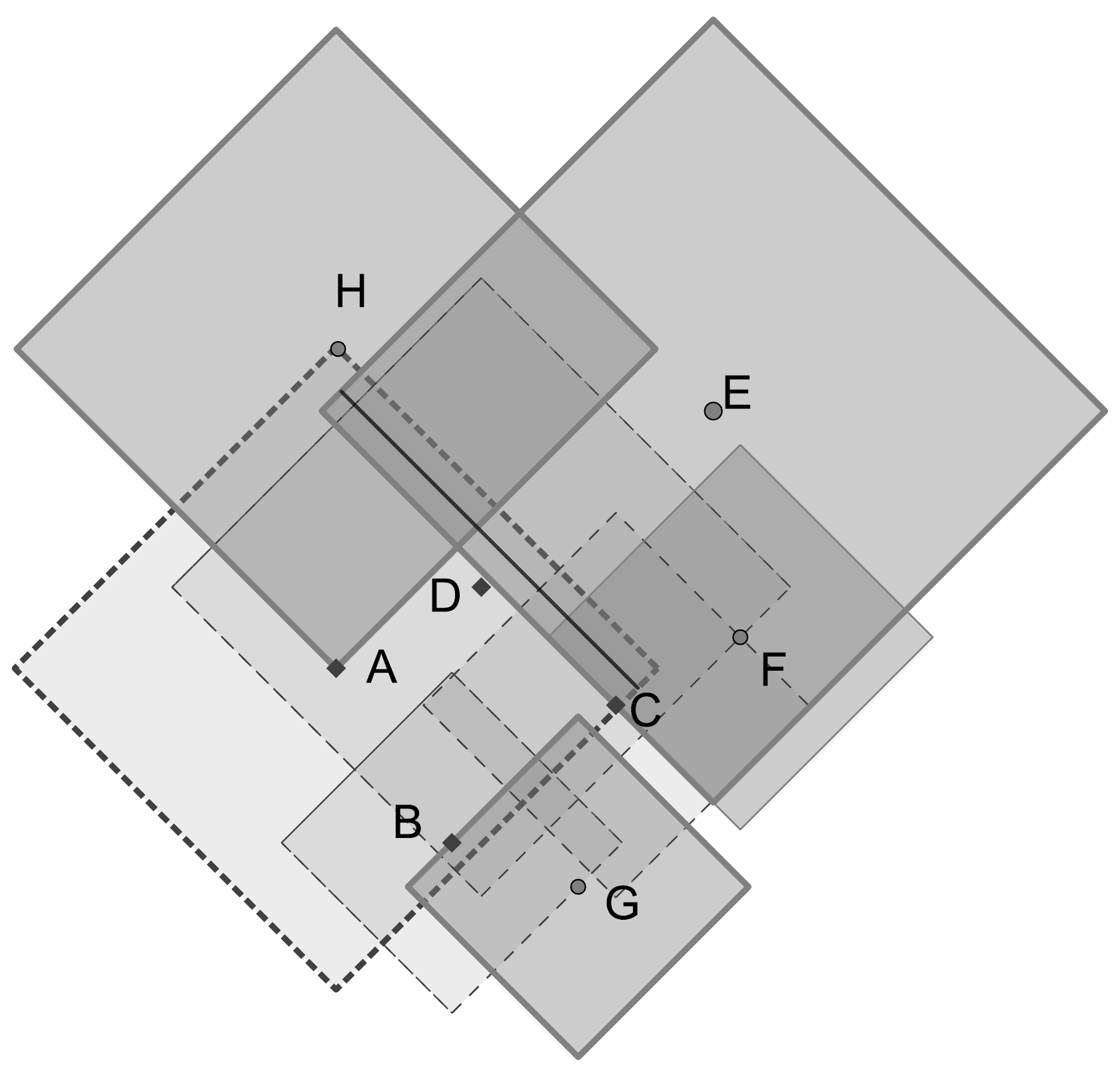}}
    \subfloat[]{\label{ejbolasc}
    \includegraphics[width=.25\linewidth]{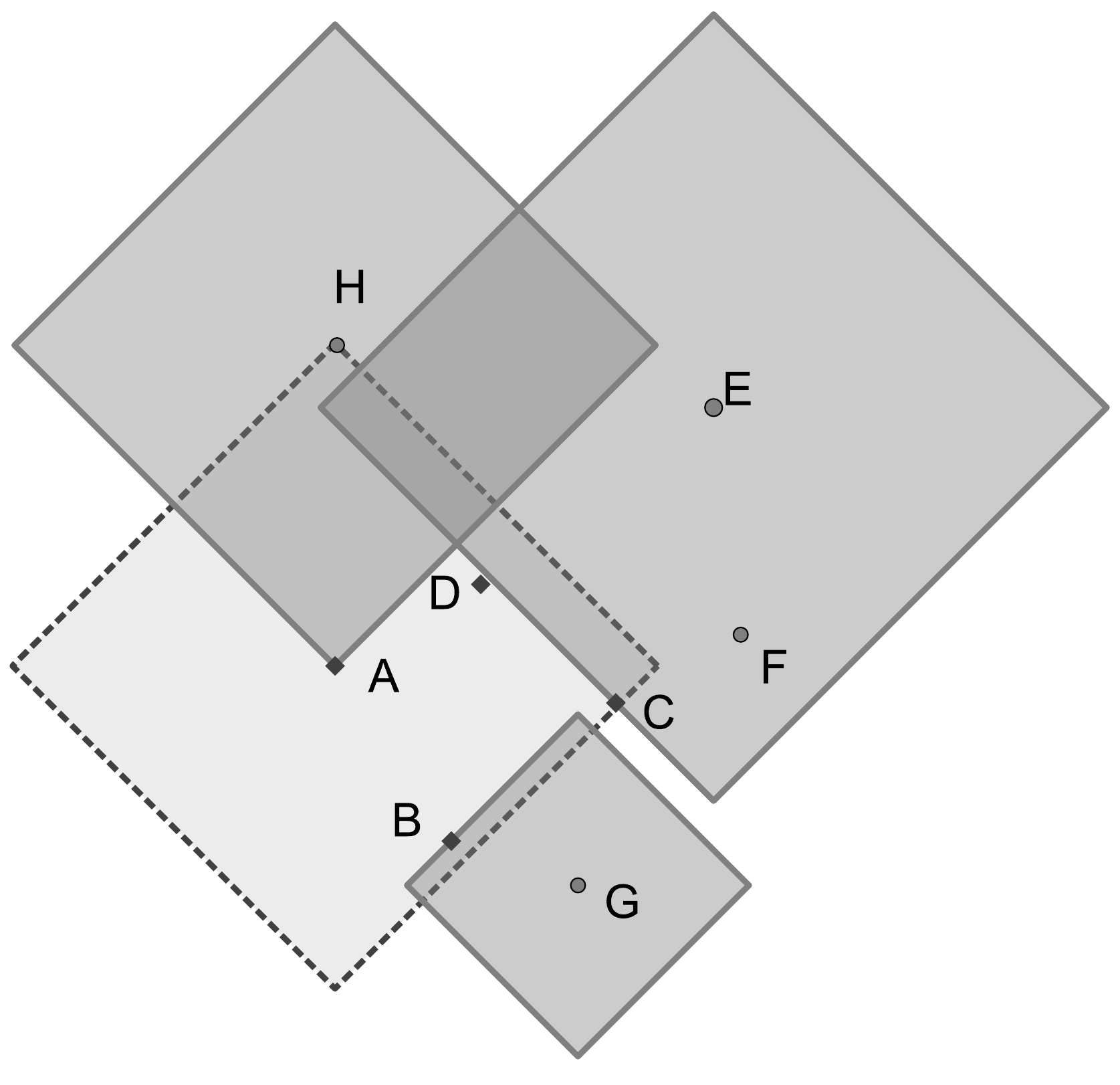}}

\caption{\label{fig:bolas}Example of the class coverage using balls generated with the Manhattan distance, where the data for each class are symbolised by circles and squares respectively. First, for all points (\ref{ejbolasa}), the balls are generated with their maximum radii that excludes instances of the opposing class (\ref{ejbolasb}); then, those including the most instances (those in thick lines) are iteratively selected until they are all covered (\ref{ejbolasc}).}
\end{figure}

Relevant data from the chosen balls, such as the central instance or the radius, are saved for further use. An example can be seen in Figure \ref{fig:bolas}. The structure of the algorithm can be seen in Algorithm \ref{alg1}, where $d_{i,j}$ is the distance between instances $i$ and $j$, $c(i)$ is the class of element $i$ (according to the classifier) and $K$ is the set of classes from the given dataset.

\begin{algorithm}[!ht]
\begin{algorithmic}[1]

\STATE  \textbf{input $d_{i,j}:i,j=1...n$; $K$}

\STATE  $D=\{\}$

\STATE  \textbf{for} $k \in K$

\STATE  $\;\;U=\{i/c(i)=k\}$

\STATE  $\;\;V=\{i/c(i)\neq k\}$

\STATE  $\;\;$\textbf{while} $U\neq\varnothing$

\STATE  $\;\;\;\;$\textbf{$\;\;$}$G=\{\}$

\STATE  $\;\;\;\;$\textbf{$\;\;$for} $i \in U$

\STATE $\;\;\;\;\;\;\;\;d=min_{j\in V}(d_{i,j})$

\STATE $\;\;\;\;\;\;\;\;P=\{u\in U/d_{i,u}<d\}$

\STATE $\;\;\;\;\;\;\;\;$\textbf{if} $|P|>|G|$

\STATE $\;\;\;\;\;\;\;\;\;\;$$G=P$

\STATE $\;\;\;\;\;\;\;\;\;\;$$l=i$

\STATE $\;\;\;\;\;\;\;\;\;\;$$r=d$

\STATE \textbf{$\;\;\;\;\;\;$end for}

\STATE  \textbf{$\;\;\;\;$}$D = D$ $\cup$ \{\{$l,c(l), r, G$\}\}

\STATE  \textbf{$\;\;\;\;U=U\setminus G$}

\STATE  $\;\;$\textbf{end while}

\STATE  \textbf{end for}

\STATE  \textbf{return }$D$

\STATE  \textbf{end}

\end{algorithmic}

\caption{Pseudocode outline of the adaptation of the P-CCCD and ONB algorithms with the maximum radius for each single-class open ball for the ONB-MACF dataset mapping.}\label{alg1}
\end{algorithm}

\subsubsection{ Instance association to the closest ball from the mapping (Algorithm \ref{alg2})}\label{subsubsec:alg2}

In this stage, the instance is associated to one of the balls from the mapping. This will help decide which part of the boundary must be used for the counterfactual generation.

Given a sample whose counterfactual should be studied, the previously generated mapping is used to associate said sample to a ball, so that the appropriate boundaries can be checked. For this, the set of balls that cover the sample is obtained (line 2). Depending on how many balls cover the instance, the association is as follows: if no ball covers the point, it is associated to the closest ball (lines 3 to 4); if only one ball covers the instance, it becomes associated to it (lines 5 to 6); if multiple balls cover the instance, it becomes associated to the ball that has the instance on its side of all boundaries with the other balls that include it (lines 7 to 15).

\begin{figure}[!ht]
\centering
\includegraphics[width=0.5\linewidth]{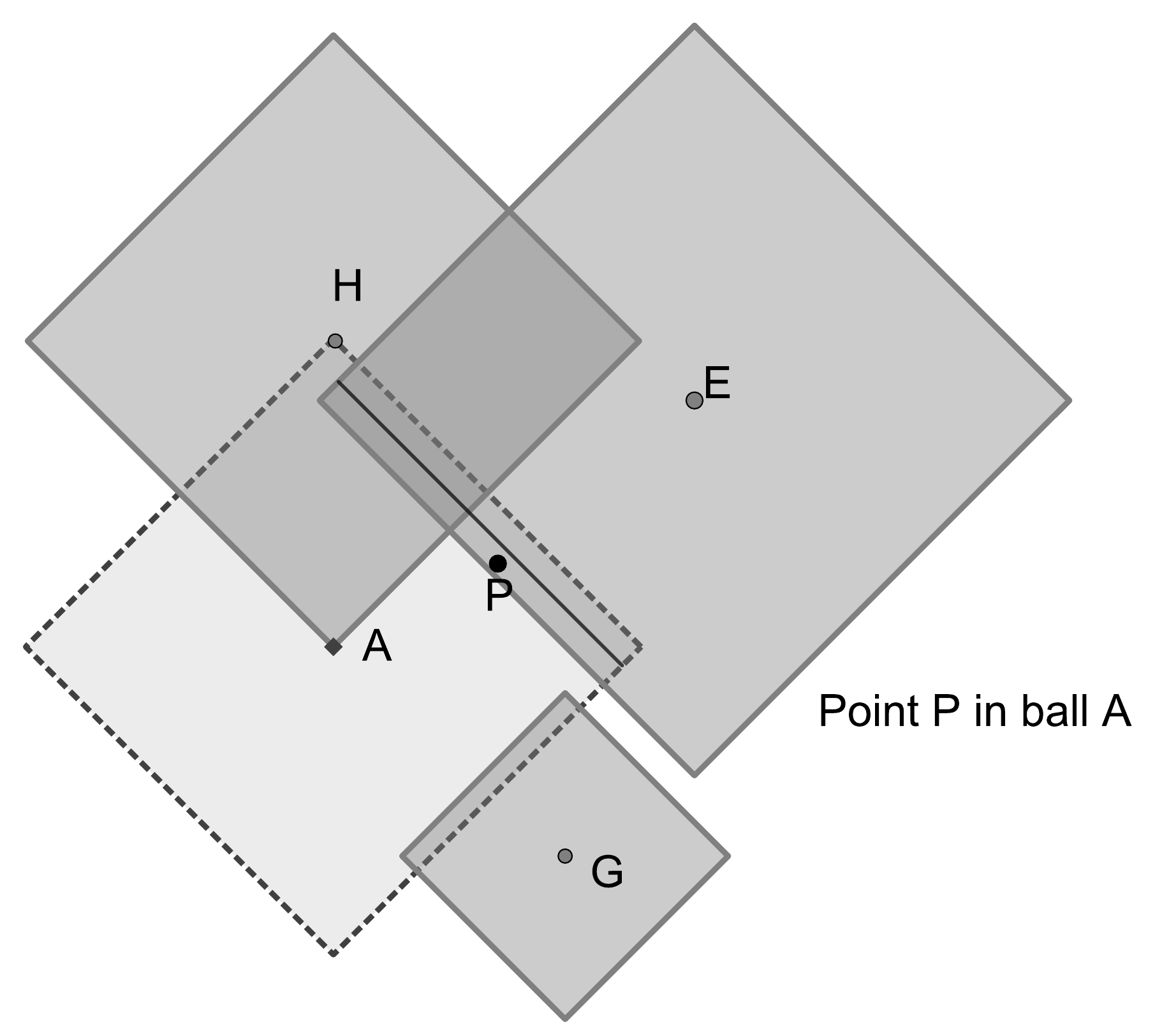}
\caption{\label{fig:fronteras}For an instance's ball association using balls generated with the Manhattan distance, the boundaries amongst balls are studied to select the most appropriate one for the studied instance. In this example, point P is inside the balls centred on A and E, and is associated to A since it falls on its side of the boundary against E.}
\end{figure}

A simple example of this association is given in Figure \ref{fig:fronteras}.

\begin{algorithm}[!ht]
\begin{algorithmic}[1]

\STATE  \textbf{input $L$; $d_{j,k}:j,k\in L$; $i$; $d_{i,l}: l \in L$}

\STATE  $E=\{l \in L : d_{i,l}<r(l)\}$

\STATE  \textbf{if} $|E|=0$

\STATE  $\;\;a = \{l \in L \mid d_{i,l} = \displaystyle\min_{l \in L} d_{i,l}\}$

\STATE  \textbf{elif} $|E|=1$

\STATE  $\;\;a = l \in E$

\STATE  \textbf{else}

\STATE  $\;\;$\textbf{for} $l \in E$

\STATE $\;\;\;\;$\textbf{if} $a = \varnothing$

\STATE $\;\;\;\;\;\;a=l$

\STATE $\;\;\;\;$\textbf{else}

\STATE $\;\;\;\;\;\;$Make the boundary between balls centred in $a$ and $l$

\STATE $\;\;\;\;\;\;$\textbf{if} $i$ falls on the side of the boundary closer to $l$

\STATE $\;\;\;\;\;\;\;\;$ $a=l$ 

\STATE \textbf{$\;\;$end for}

\STATE  \textbf{return }$a$

\STATE  \textbf{end}

\end{algorithmic}

\caption{Pseudocode outline of the ball association process for instance $i$, where $L$ is the set of balls from Algorithm \ref{alg1}, with each ball centred in $l$, and where $r(l)$ and $G(l)$ are the radius and the elements covered by the ball centred in l respectively.}\label{alg2}
\end{algorithm}

\subsubsection{ Counterfactual candidate generation (Algorithm \ref{alg3})}\label{subsubsec:alg3}
 
This is the stage where counterfactuals are generated for the given instance. The process proceeds as follows. If the assigned ball shares the class with the studied instance, the closest ball from an opposing class is searched for using the mapping (lines 4 to 10), for which the boundaries between the assigned ball and the opposing ones are calculated. 

In case there are immutable features, instead of the opposing balls and their centres, their projections on the subspace where said features are restricted to having the same values as the studied instance are used instead for this step (viable projections are inside their balls and maintain the ball's class) so that all candidates will share those immutable values (line 5). 

Then, the absolute differences in each feature between the sample and the projected centre are ordered and, starting with the feature with the lowest difference, some features of the latter are modified to take the instance's values (if they keep the class intact) to improve the counterfactual's sparsity (line 6).

\begin{figure}[!ht]
\centering
\includegraphics[width=0.8\linewidth]{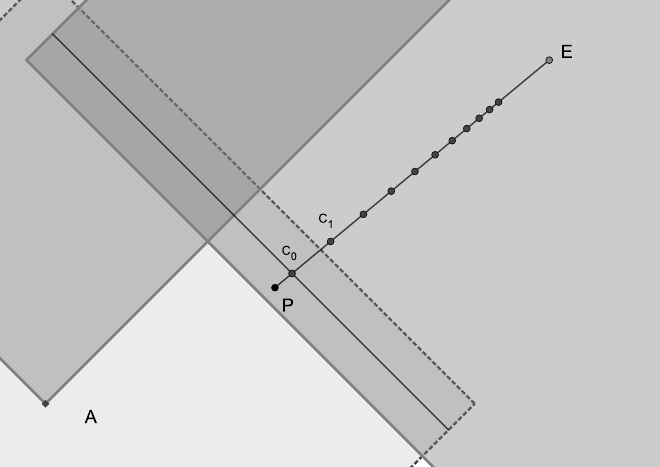}
\caption{\label{fig:candidatos}For an instance's candidate counterfactual generation with a chosen opposing ball in a dataset with no immutable nor discrete features, a first candidate $C_{0}$ is given at the intersection of the boundary between the instance P's associated ball and the chosen opposing ball E and the segment that goes from the instance to the centre of the selected opposing ball. If that candidate's class is the given instance's, subsequent candidates $C_{i}$ are generated in the said segment in intervals with decreasing length until the class changes or a check limit is reached (in which case the counterfactual would be E).}
\end{figure}

After finding the candidate balls with the closest boundaries and valid projected centres (that stay inside the balls and keep the class), for each chosen ball, a first candidate is created at the intersection of the boundary and the straight line connecting the sample and the projected centre (line 7). If the class is not the opposing ball's, further counterfactual candidates are sequentially created on that straight line in decreasing intervals towards the projected centre until a counterfactual with the target class is reached (lines 21 to 33, and an example of this is included in Figure \ref{fig:candidatos}). If this process did not find a counterfactual immediately, the last candidate with the same class as the original instance would be a valid semifactual.

If no candidates are found, the process is repeated after relaxing the condition for viable projections to just having the correct class (lines 11 to 12) and, if the process provides no candidates yet again, the immutability restrictions are withheld (lines 13 to 14). As before, from each boundary candidate, further candidates are obtained towards their associated projections in decreasing intervals until the class changes to the ball's. 

Had the initially associated ball not shared its class with the instance, the dictionary would be updated by reorganising the ball so that it does not include the studied instance (which is done by using Algorithm \ref{alg1} on the set of data points covered by the ball, including the new instance, as in lines 15 to 20), and the process is the same as earlier from there. 

If there are discrete features, their values on the candidates are always rounded to the nearest integer throughout the process to maintain such property.

\begin{algorithm}[!ht]
\footnotesize
\begin{algorithmic}[1]

\STATE  \textbf{input $D$; $L$; $d_{j,k}:j,k\in L$; $i$; $a$; $d_{i,l}: l \in L$; $C_{target}$; $n$; $r$}

\STATE  BCL$=\{\}$

\STATE  \textbf{if} $c(a)=c(i)$

\STATE  $\;\;$\textbf{for} $l \in L_{|_{C_{target}}}$

\STATE $\;\;\;\;$\textbf{if} the projection of $l$ is viable

\STATE $\;\;\;\;\;\;$Reduce changes between said projection and $i$

\STATE $\;\;\;\;\;\;$Take cf candidate on the boundary towards it

\STATE $\;\;\;\;\;\;$\textbf{if} $d_{i, cand}<d_{i,j}$ for $j \in BCL$ or $|BCL|=0$

\STATE $\;\;\;\;\;\;\;\;$ update $BCL$ with $cand$ (keeping $|BCL|\leq n$)

\STATE \textbf{$\;\;$end for}

\STATE  $\;\;$\textbf{if} $|BCL|=0$

\STATE  $\;\;\;\;$Same as 3-10, relaxing the projection validity restriction 

\STATE  $\;\;$\textbf{if} $|BCL|=0$

\STATE  $\;\;\;\;$Same as 3-10 also withholding immutability restrictions

\STATE  \textbf{else}

\STATE  $\;\;$Repeat Algorithm \ref{alg1} on $G(a) \cup \{i\}$, obtaining $D'$

\STATE  $\;\;D=D \setminus \{a, c(a), r(a), G(a)\}$

\STATE  $\;\;$$D=D U D'$

\STATE  $\;\;$$a=i$

\STATE  $\;\;$Same as 3-14 with the updated $D$

\STATE  \textbf{for} $cand$ in $BCL$

\STATE  $\;\;$$found$ = $false$

\STATE  $\;\;t=0$

\STATE  $\;\;V$ is the vector from $cand$ to the associated projection

\STATE  \textbf{$\;\;$while} $found$ = $false$ and $t<10$

\STATE  $\;\;\;\;$$cand=cand+r^tV$

\STATE  $\;\;\;\;$\textbf{if} c($cand$)=c($proj$)

\STATE  $\;\;\;\;\;\;$$found$ = $true$

\STATE  $\;\;\;\;$$t=t+1$

\STATE  \textbf{$\;\;$end while}

\STATE  $\;\;$\textbf{if} $found$=$false$

\STATE  $\;\;\;\;cand$=$proj$

\STATE  $\;\;$save $cand$ as cf on list of counterfactuals for $i$

\STATE  \textbf{end for}

\STATE  \textbf{return }list of counterfactuals for $i$

\STATE  \textbf{end}

\end{algorithmic}

\caption{Pseudocode outline of the counterfactual generation process for instance $i$, where $D$ and $L$ are the dictionary and set of balls from Algorithm \ref{alg1} respectively, with each ball centred in $l$, $a$ is the associated ball for $i$ from Algorithm \ref{alg2}, $C_{target}$ is the set of target classes for instance $i$, $n$ is the number of counterfactuals to be given, $r$ is the ratio for the decreasing intervals and $BCL$ is the best candidate list.}\label{alg3}
\end{algorithm}

\subsection{Advantages of the ONB-MACF method}\label{advantages}

The design of the proposed ONB-MACF method poses multiple advantages. Due to being model-agnostic, it only requires access to the prediction function from the model, furthering its versatility. While it was designed for tabular data, it might be able to be used on text or image datasets after some data transformations \cite{de_oliveira_model-agnostic_2023}. Furthermore, unlike other methods whose strategies involve random modifications of the instance's features \cite{laugel_comparison-based_2018}  or swapping feature values \cite{brughmans_nice_2023}, ONB-MACF yields deterministic results. The identification and utilisation of class boundaries instead of simply other instances also allows for closer counterfactuals. Additionally, unlike most methods, ONB-MACF can propose semifactuals, further bolstering explainability.

The methodology followed for the generation of counterfactuals by the ONB-MACF method also has several advantages.

Firstly, it must be noted that there is a single computationally intensive step, namely the generation of the dataset mapping (Algorithm \ref{alg1}). 
Considering the usefulness of this process, we may highlight two clear advantages associated with this computation:

\begin{itemize}
    \item for repeated use of a dataset and its associated classification model, which is very common in real-life situations where the outcome of the model itself needs to be audited, this step only has to be performed once, thus avoiding unnecessary computations, and
    \item the mapping reduces the search space size for other stages (from the number of dataset instances to the number of balls in the coverage). Thus, fewer comparisons are necessary for counterfactual generation than other instance-based counterfactual strategies \cite{poyiadzi_face_2020}.
\end{itemize}

Secondly, for the generation of counterfactual candidates, the procedure stated in Algorithm \ref{alg3} includes several features that allow it to be customised to different cases:

\begin{itemize}
    \item counterfactuals for multiple samples can be obtained in a single run;
    \item for each sample, a target number of counterfactuals can be generated, using different opposing balls for each case in order to foster better counterfactual diversity;
    \item in multiclass cases, target counterfactual classes can be specified;
    \item immutability of features is maintained; and 
    \item discrete and categorical features maintain their properties.
\end{itemize}

\section{Experimental Framework}\label{frame}

To analyse the sound capabilities of the proposed ONB-MACF method, in this section, we provide the required material for a fair comparison with respect to the state of the art. In order to ease the compilation of the necessary resources, the CARLA\footnote{CARLA Benchmark, \url{https://github.com/carla-recourse/CARLA}, last accessed 2 April 2024} benchmarking package \cite{pawelczyk_carla_2021} was used. It includes implementations of counterfactual explanation and recourse methods from literature, as well as datasets, pre-trained models and performance metrics that evaluate the methods from different perspectives. For further proof of the capabilities of the ONB-MACF method, a qualitative analysis of the generated counterfactuals compared to those of Growing Spheres and NICE was performed. 

This experimental framework section comprises the following structure. Firstly, a short description of the datasets is included in Section \ref{datasets}. Secondly, the structure of the classifier for each dataset and the different methods against which our proposal will be tested are indicated in Section \ref{classifier}. 
Finally, the metrics indicating how their performance will be measured are detailed in Section \ref{metrics}.

\subsection{Datasets}\label{datasets}

This subsection describes the characteristics of the datasets used in the experimental study. 

The benchmarking study includes eight well-known binary datasets: 
``adult'', ``COMPAS'', ``Give Me Some Credit'', ``HELOC'', ``Irish'', ``Saheart'', ``Titanic'' and ``Wine''. 

\begin{itemize}

\item The ``adult'' dataset evaluates whether a person will earn over 50,000 dollars a year according to census data. While this dataset initially included 48,832 instances, its size was halved to 24,416 samples by stratified undersampling to reduce execution times while maintaining the structure and representativity of the dataset. It contains 13 features and the binary class, of which 7 are categorical (binary) attributes and 2 are considered immutable (``age'' and ``sex'').

\item The ``COMPAS'' dataset predicts whether convicts will re-offend in the following two years based on their personal data and criminal record. It includes 6,172 samples, with 7 features and the binary class, of which 3 are categorical (binary) and 3 are immutable (``age'', ``race'', ``sex'').

\item The ``Give Me Some Credit'' dataset evaluates whether a person will have financial problems in the next two years, to give or deny them a loan. While originally including 115,527 instances, its size was reduced to a fifth by stratified undersampling (23,105) to reduce execution times while maintaining the structure and representativity of the dataset. It has 10 variables and the binary class, all of which take continuous values and where only 1 feature is immutable (``age'').

\item The ``HELOC'' dataset is a credit dataset that predicts whether people will be able to pay their loans in two years. It includes 9871 samples, with 21 features (all of them numeric and none of them immutable) and the binary class.

\item The ``Irish'' dataset evaluates whether Irish schoolchildren aged 11 in 1967 would take their Leaving Certificate of their studies (which implied passing their final examinations). This dataset includes 500 samples, with 5 features and the binary class, of which 3 are categorical (1 is binary, 2 are one-hot-encoded due to having multiple options) and 1 is immutable (``sex'').

\item The ``Saheart'' dataset predicts whether males in a high-risk heart-disease region of the Western Cape in South Africa have a coronary heart disease. This dataset includes 462 samples, 9 features and the binary class, of which 1 is categorical (binary) and 1 is immutable (``age'').

\item The ``Titanic'' dataset includes data from the passengers of the Titanic and whether they survived the accident. It includes 2099 instances, 8 features and the binary class, of which 4 are categorical (1 is binary, 3 are one-hot-encoded due to their multiple options) and 3 are immutable (``gender'', ``age'' and ``country'').

\item The ``Wine'' (``Wine quality white'') dataset evaluates the quality of Portuguese white wines according to their chemical composition and the grades given by experts. While this dataset is often used for regression, as the grades are given out of 10, the target was transformed to binary (passing grade or not). It includes 4898 instances, 11 attributes, all of them numeric, and the binary class. 
\end{itemize}

From each dataset, a sample of 200 instances (enough to provide some variety of inputs) will be separated. Their classes will be predicted using the benchmarking classifier and they will serve as factuals whose predictions want to be explained.

\subsection{Classifier configuration and explanation methods}\label{classifier}

Counterfactual methods aim to explain and give further insight into the decisions of a classifier. The chosen model was a neural network because explanation methods are usually applied to black-box models. 

For each dataset, a pre-trained neural network with two hidden fully-connected layers with 32 and 16 neurons respectively, and a ReLU activation function was used, 
akin to CARLA's benchmarking. All counterfactual algorithms must work on the same model for fair comparison on each dataset.

Regarding the counterfactual explanation algorithms from the state of the art, CEM \cite{dhurandhar_explanations_2018} (on its base and variational autoencoders versions), CLUE \cite{antoran_getting_2021}, CRUDS \cite{downs_cruds_nodate}, DiCE \cite{mothilal_explaining_2020}, FACE \cite{poyiadzi_face_2020} (on its knn and $\epsilon$-graph versions), Growing Spheres \cite{laugel_comparison-based_2018}, Wachter \cite{wachter_counterfactual_2017} and NICE \cite{brughmans_nice_2023}, whose particularities were explained in Section \ref{subsec:related}, were included in the quantitative benchmarking.

\subsection{Counterfactual performance metrics and evaluation approach}\label{metrics}
We evaluate the performance of the generated counterfactuals using several metrics, which are enumerated below.
Each of these metrics is related to one or more properties of interest for counterfactuals, as described in Section \ref{basics}.
\begin{itemize}
    \item \textbf{L0 norm}: it measures the L0 norm between the studied sample and its counterfactual; that is, how many attributes change between them. It measures \textbf{sparsity}.
    \item \textbf{L1 norm}: it measures the Manhattan distance between the studied sample and its counterfactual, which  evaluates \textbf{similarity/closeness}.
    \item \textbf{L2 norm}: it measures the squared Euclidean distance between the studied sample and its counterfactual, which also gauges \textbf{similarity/closeness}.
    \item \textbf{L$\infty$ norm}: it measures the L$\infty$-norm between the studied sample and its counterfactual, which evaluates \textbf{similarity/closeness}, too.
    \item \textbf{Constraint violation}: it measures how many immutability restrictions have failed in an explanation. It is a metric of \textbf{actionability}.
    \item \textbf{Redundancy}: it checks how many of the changed attributes between the sample and its counterfactual could be reverted without modifying the counterfactual's class (that is, how many changes were unnecessary). This also measures \textbf{sparsity}.
    \item \textbf{Y-NN}: it checks the ratio of same-class elements amongst the $Y$ (5) nearest neighbours of each counterfactual, which measures \textbf{plausibility} but is related to \textbf{similarity}.
    \item \textbf{Success rate}: it indicates the ratio of samples whose counterfactuals were successfully found, which is a matter of \textbf{validity} of counterfactuals.

\end{itemize}

The presented metric for the evaluation of each method is the arithmetic mean of the results for all the studied samples of each dataset. By default, the y-NN and success rate metrics are better for values closer to 1, while the rest are better for values closer to 0. It must also be noted that the metrics from CARLA are unable to evaluate counterfactual diversity (possibly due to not many methods are explicitly designed to produce different counterfactuals over which diversity can be measured) and, even though the L1, L2 and L$\infty$ norms measure the same properties, they do so from different points of view: the Manhattan distance can correlate to the Euclidean one (but not always, depending on the distribution of feature changes), and the L$\infty$ norm gives information on the extremeness of changes in individual features.

\section{Quantitative performance analysis of the ONB-MACF method}\label{exper}

This section shows the results obtained with each dataset and 
counterfactual method, carrying out a thorough study 
of the good behaviour of the ONB-MACF method. 
The results are evaluated following the characteristics of their generated counterfactuals according to the metrics enumerated in Section \ref{metrics}.

For each dataset, the counterfactual explainers from Section 
\ref{classifier} are compared to the proposed ONB-MACF method. 
Given the differences in dataset characteristics, the metrics are obtained and transformed separately for each dataset, so that all metrics go from 0 to 1 and higher values indicate better performance. The results are then aggregated. 
While the goodness of a method according to a particular metric can be easily seen, its general goodness is evidenced by the mean of the metrics. These means are presented in Table \ref{tab_average_results}, along with the averages of the results (scaled per dataset) for each metric and method.  
The raw metric results per dataset are presented in Tables \ref{tab_adult} to \ref{tab_wine} of \ref{result_tables}.

\begin{figure}[!ht]
\centering

    \subfloat[][``Adult'']{\label{fig_adult}
    \includegraphics[width=0.37\linewidth]{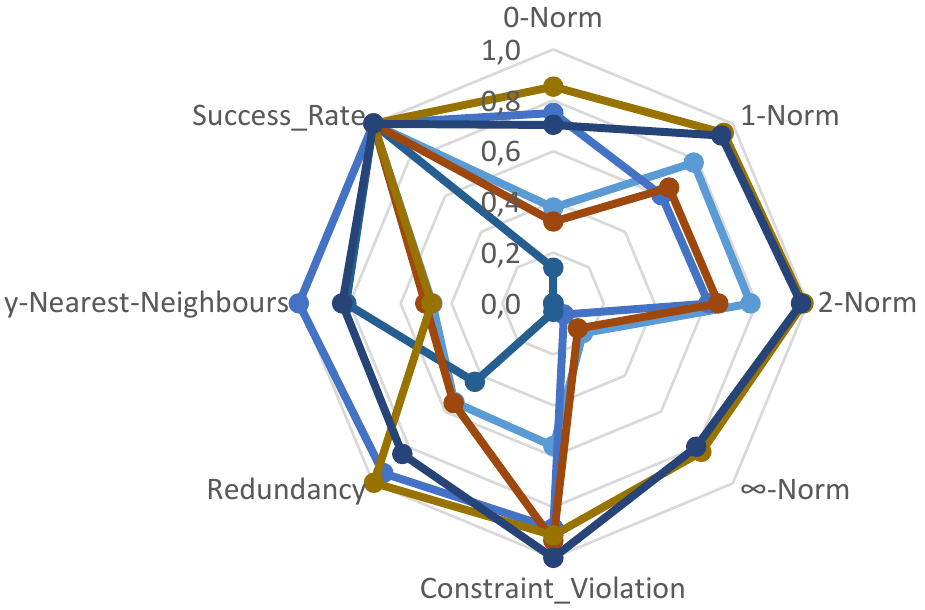} }
    \hfil
    \subfloat[][``COMPAS'']{\label{fig_compas}
    \includegraphics[width=0.37\linewidth]{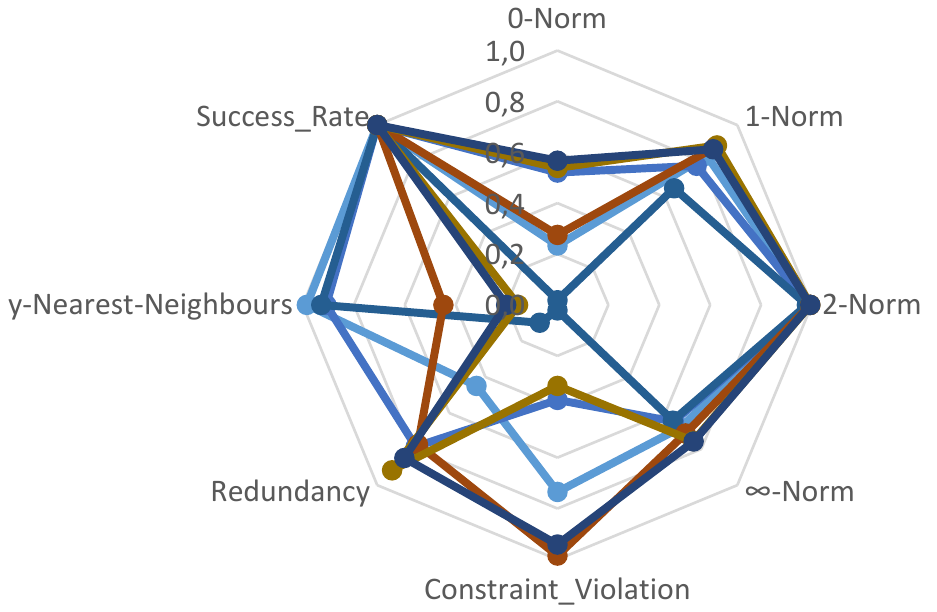}}
    \hfil
    \subfloat[][``Give Me Some Credit'']{\label{fig_give}
    \includegraphics[width=0.37\linewidth]{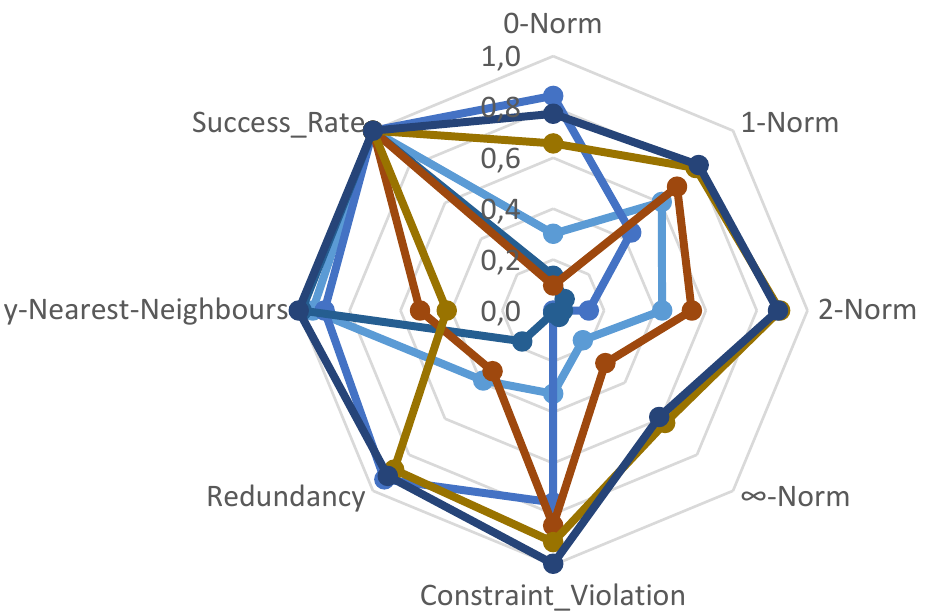}}
    \hfil
    \subfloat[][``HELOC'']{\label{fig_heloc}
    \includegraphics[width=0.37\linewidth]{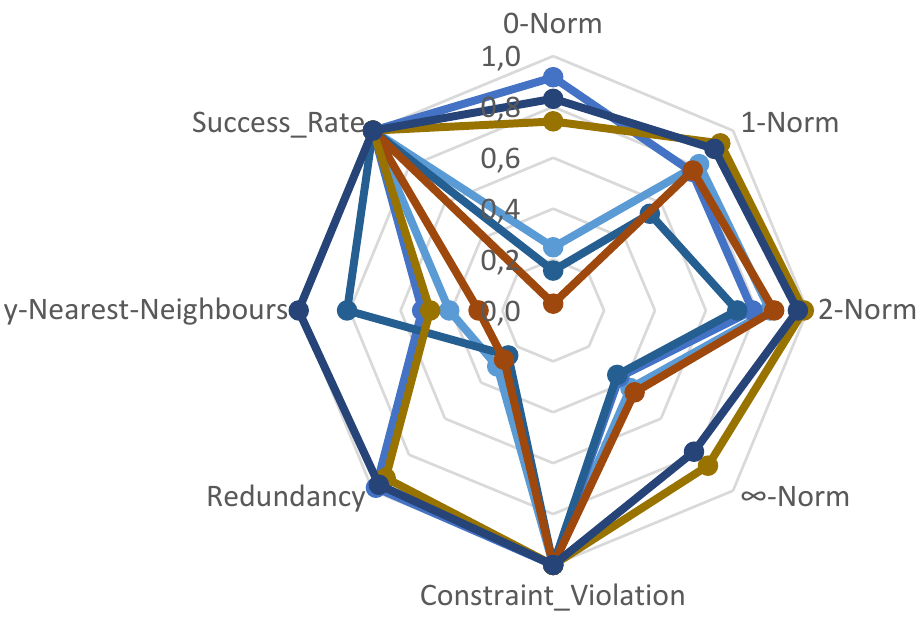}}
    \hfil
    \subfloat[][``Irish'']{\label{fig_irish}
    \includegraphics[width=0.37\linewidth]{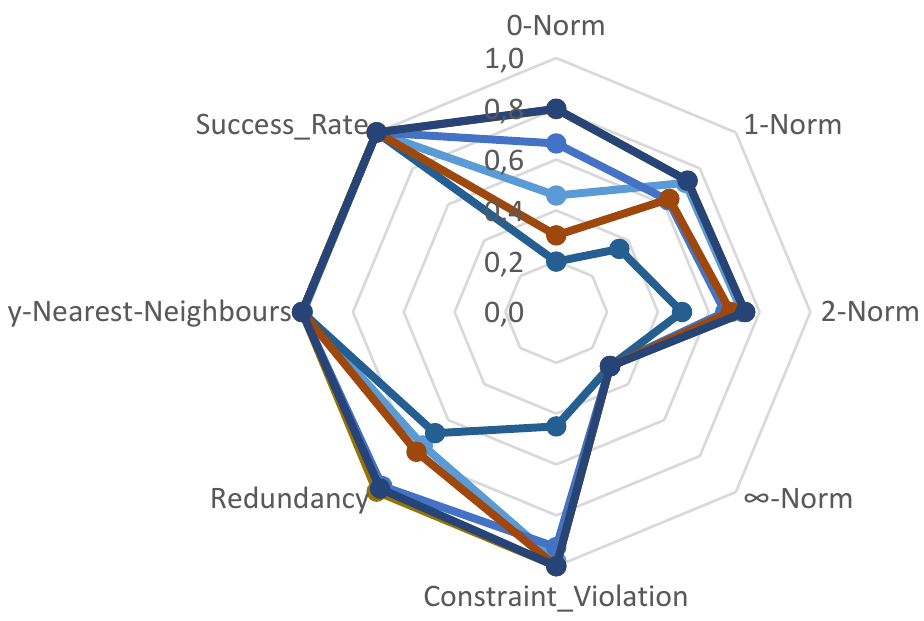}}
    \hfil
    \subfloat[][``Saheart'']{\label{fig_saheart}
    \includegraphics[width=0.37\linewidth]{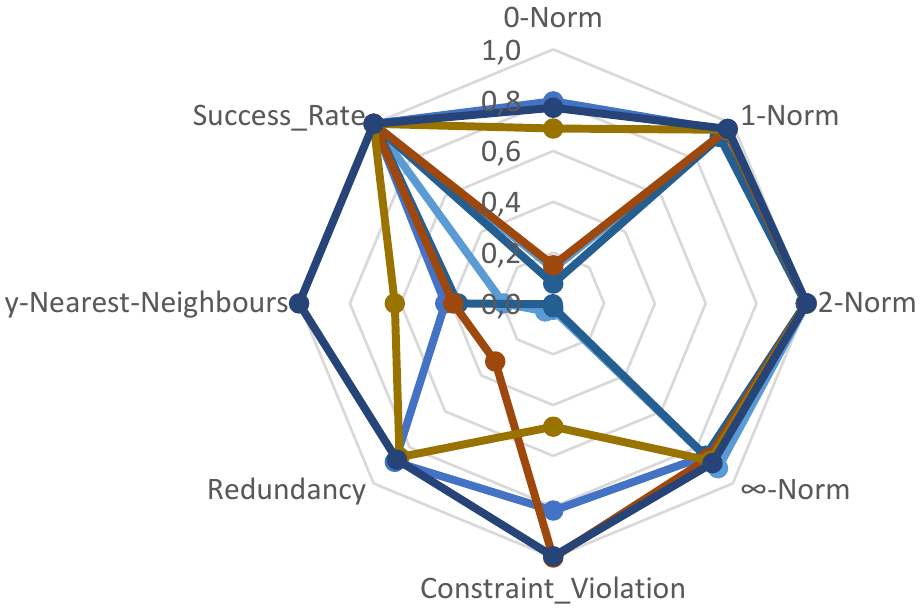}}
    \hfil
    \subfloat[][``Titanic'']{\label{fig_titanic}
    \includegraphics[width=0.37\linewidth]{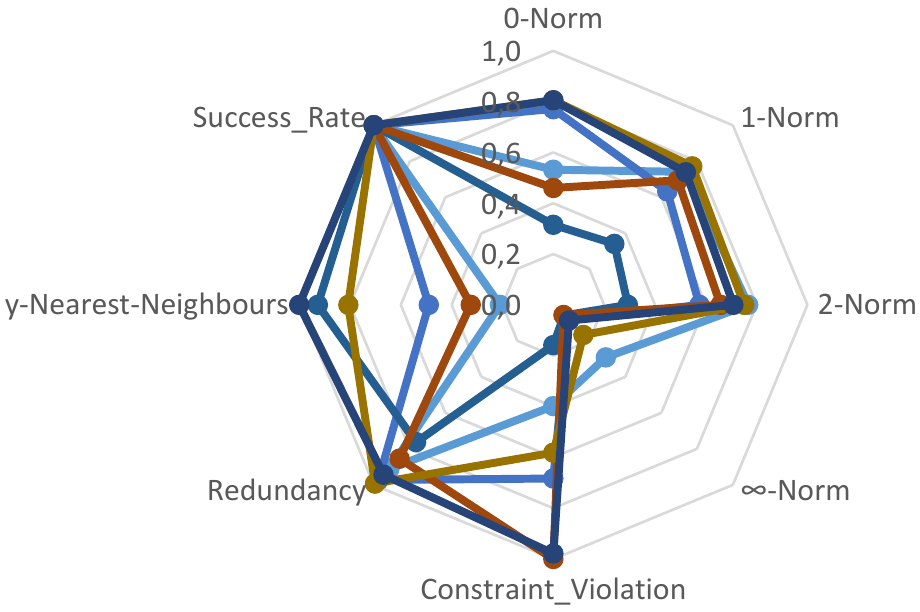}}
    \hfil
    \subfloat[][``Wine'']{\label{fig_wine}
    \includegraphics[width=0.37\linewidth]{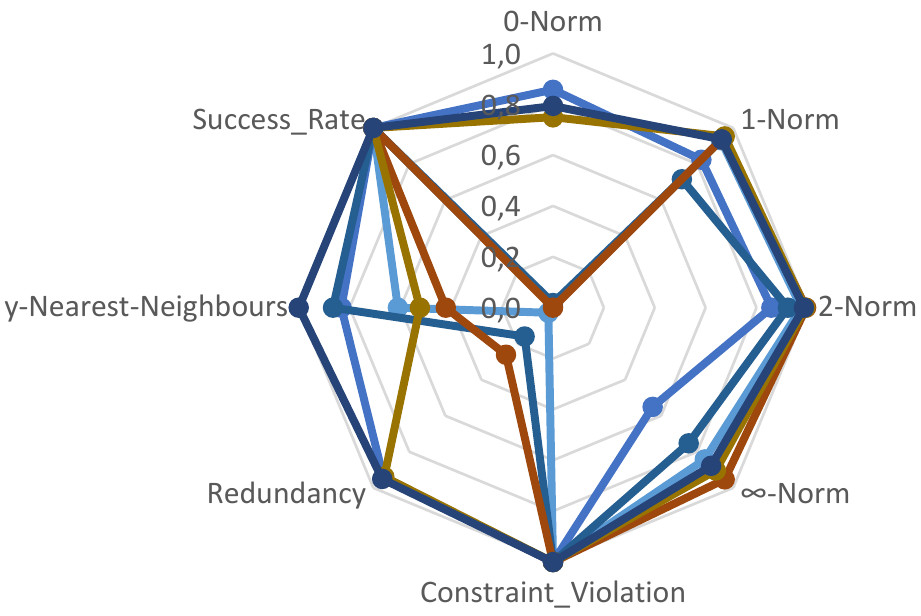}} 
    \hfil    
    \subfloat{\label{legend}
    \includegraphics[width=0.74\linewidth]{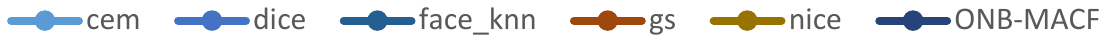}}

\caption{\label{fig:many_starplots}Comparison of the best-performing methods for the ``Adult'' (\ref{fig_adult}), ``COMPAS'' (\ref{fig_compas}), ``Give Me Some Credit'' (\ref{fig_give}), ``HELOC'' (\ref{fig_heloc}), ``Irish'' (\ref{fig_irish}), ``Saheart'' (\ref{fig_saheart}), ``Titanic'' (\ref{fig_titanic}) and ``Wine'' (\ref{fig_wine}) datasets.}
\end{figure}

\begin{figure}[!h]
\centerline{\includegraphics[width=0.9\linewidth]{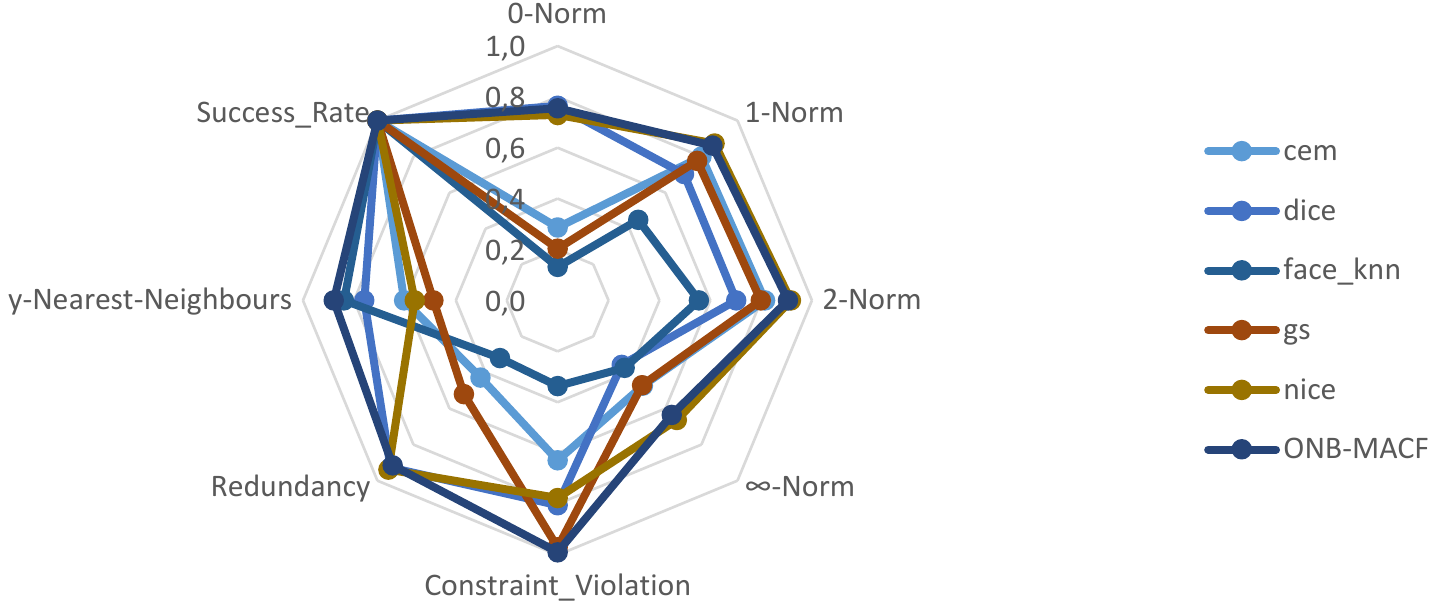}}
\caption{Comparison of the best performing methods, averaged over the datasets.}
\label{fig_general}
\end{figure}

For the sake of a better understanding, results are graphically summarized in the form of radial plots. In these radial plots, all metrics from Section \ref{metrics} are associated with the vertices of a regular polygon, following the same order from that list in a clockwise fashion. For radial plot generation, and similarly to Table \ref{tab_average_results}, values in all metrics were transformed so that the scale went from 0 to 1 and so higher values indicated better performance. 
Each method's resulting scaled metric values are located in the segment between the centre and that metric's vertex and joined into another polygon. Results for every individual dataset are shown in Figure \ref{fig:many_starplots}, while the averaged comparison of the best-performing methods can be found in Figure \ref{fig_general}.

\begin{table*}[htbp]
\caption{Average of the performance results, scaled per dataset, with an overall mean over the metrics.}
\begin{center}
\scalebox{0.7}{
\begin{tabular}{cccccccccc}
\toprule
\textbf{Method}&\textbf{L0 norm}&\textbf{L1 norm}&\textbf{L2 norm}&\textbf{L-$\infty$ norm}&\textbf{Const. Vio.}&\textbf{Redun.}&\textbf{YNN}&\textbf{Succ. Rate}&\textbf{Overall mean} \\

\toprule
CEM & 0.287 & 0.799 & 0.815 & 0.474 & 0.628 & 0.428 & 0.603 & \textbf{1.000} & 0.629 \\

CEM-VAE & 0.286 & 0.798 & 0.814 & 0.474 & 0.623 & 0.430 & 0.606 & \textbf{1.000} & 0.629 \\

CLUE & 0.019 & 0.138 & 0.252 & 0.178 & 0.353 & 0.032 & 0.713 & 0.663 & 0.293 \\

CRUDS & 0.015 & 0.259 & 0.354 & 0.229 & 0.395 & 0.045 & 0.886 & 0.641 & 0.353 \\

DICE & 0.765 & 0.703 & 0.702 & 0.357 & 0.804 & 0.928 & 0.759 & \textbf{1.000} & 0.752 \\

FACE-EPSILON & 0.131 & 0.445 & 0.553 & 0.375 & 0.324 & 0.315 & 0.849 & 0.946 & 0.492 \\

FACE-KNN & 0.133 & 0.448 & 0.556 & 0.374 & 0.336 & 0.319 & 0.838 & 0.999 & 0.501 \\

GS & 0.205 & 0.777 & 0.799 & 0.471 & 0.970 & 0.520 & 0.488 & \textbf{1.000} & 0.654 \\

WACHTER & 0.082 & 0.474 & 0.512 & 0.305 & 0.424 & 0.480 & 0.450 & 0.860 & 0.448 \\
NICE & 0.729 & \textbf{0.873} & \textbf{0.917} & \textbf{0.664} & 0.776 & \textbf{0.939} & 0.561 & \textbf{1.000} & 0.808 \\
\hline
ONB-MACF & \textbf{0.756} & 0.861 & 0.906 & 0.635 & \textbf{0.989} & 0.915 & \textbf{0.879} & \textbf{1.000} & \textbf{0.868} \\
\hline

\end{tabular}}
\label{tab_average_results}
\end{center}
\end{table*}

To properly evaluate the general quality of the counterfactuals obtained using the ONB-MACF method, the results from the 8 cases of study will be summarised 
from the multiple fronts recommended in Section \ref{basics} and reflected by the metrics.

\begin{itemize}
    \item \textbf{Sparsity}. The ONB-MACF method consistently achieves good results, as measured by the L0 norm, with changes in generally 1-3 features as recommended, so the 
    counterfactuals should be easy for humans to grasp. This point is further proven by the low (sparsity-related) redundancy values.
    \item \textbf{Similarity}. Evaluated using the L1, L2 and L-$\infty$ norms, the ONB-MACF method 
    presents excellent results in this regard, with close counterfactuals to the given instances and without lopsided 
    feature changes 
    (evidenced by low L-$\infty$ norm values).
    \item \textbf{Actionability}. The method obtains 
    very few constraint violations, which is 
    highly positive. 
    The immutability restrictions are always followed for an instance unless there are no suitable coverage balls (that is, there are no balls whose centre's projection onto the values of the immutable features of the instance maintains that ball's class).
    \item \textbf{Plausibility}. The YNN values show that the ONB-MACF method can achieve some of the best results on datasets without a substantial percentage of immutable and discrete features.
    \item \textbf{Validity}. As measured by the Success Rate, the method found counterfactuals for all samples, just like most other studied counterfactual explainers.
    \item \textbf{Diversity}. While the metrics from CARLA cannot evaluate this property, the ONB-MACF method 
    promotes it. When 
    multiple counterfactuals are required for a sample, the counterfactual candidates are generated using different balls. 
    Thus, counterfactuals are created in different directions and using different combinations of features.
\end{itemize}

It must be noted that COMPAS seems to be a problematic dataset to evaluate since almost half of the features are immutable and categorical, which, combined with the lower number of samples, makes it challenging to find feasible candidates that do not violate any constraints. On the other hand, Give Me Some Credit is the type of dataset where the ONB-MACF method would be recommended, as the lack of categorical variables makes it easier for the algorithm to find candidates using straight lines to the projected centres of opposing balls and to optimise the sparsity of the final counterfactuals.

On another note, regarding computational speed, the ONB-MACF method is generally comparable to the other slow counterfactual methods when the distance matrix and the dictionary have been calculated beforehand (as they only need to be obtained the first time a dataset is evaluated and can be saved). The speed is adequate on curated datasets with limited data to be used. A similarly scaled ranking of the time results can be seen in Table \ref{avg_time_tab} of \ref{result_tables}.

\section{Qualitative analysis of the ONB-MACF method}\label{qualitative}

This section qualitatively assesses the capabilities of the ONB-MACF method. For this qualitative analysis
, 200 counterfactuals were obtained on the neural network classifiers for each of the four chosen datasets (``Irish'', ``Saheart'', ``Titanic'' and ``Wine quality white''). The counterfactuals given by the ONB-MACF method are scrutinised and compared to those obtained with two similar methods that share similarities with it:

\begin{itemize}
    \item Growing Spheres \cite{laugel_comparison-based_2018} uses hyperspheres centred on the instances, which may seem similar to the ONB-MACF method. However, the use of multiple balls simultaneously in our proposal and the better exploitation of their morphology, compared to the random instance generation in growing spheres, sets both methods apart.
    \item NICE \cite{brughmans_nice_2023} is an instance-based counterfactual method that uses opposite-class members as prototypes, like the ONB-MACF method. NICE's strategy is to iteratively swap feature values of the instance to explain and those of the dataset's prototype, while the ONB-MACF method goes a step beyond by generating its own candidates.
\end{itemize}

The differences amongst counterfactual strategies can lead to very different types of counterfactuals, which can affect the auditability of the classification models. In broad terms, counterfactuals can be used to check whether the recommended changes make sense in particular cases, but when you generate enough of them, you can also aggregate the results and see which variables are generally causing the class change.

In this qualitative study, the focus will be on four questions:

\begin{itemize}
    \item which features are most often used for class change,
    \item which features are often used for class change together,
    \item wether the change in class according to those features would make sense in the eyes of an expert, and
    \item whether the features most used for class change differ amongst counterfactual methods.
\end{itemize}

The first part of Table \ref{tab_all_qual} shows the feature changes on the counterfactuals for the Irish dataset. In this dataset, the most important feature is Education level, as all methods suggest. In this case, both ONB-MACF and NICE properly point to changes in that feature, while Growing spheres also includes changes in most other non-immutable features. Thus, the auditability according to ONB-MACF and NICE is very good (the proper feature is the one used), and the quality is diluted while using GS.

\begin{table*}[htbp]
\footnotesize
\caption{Number of changes per feature on each dataset for each counterfactual method.}
\begin{center}
\scalebox{0.8}{
\begin{tabular}{ccccccc|c}
\toprule

\multirow{2}{*}{\textbf{Dataset}} & \multirow{2}{*}{\textbf{Feature}} & \multicolumn{3}{c}{\textbf{Feature changes per method}}\\
  &   & \textbf{GS} & \textbf{NICE} & \textbf{ONB-MACF} \\
\toprule
\multirow{6}{*}{Irish}& Sex & 0 & 0 & 0 \\
& DVRT & 200 & 0 & 0 \\
& Education level & 200 & 200 & 200 \\
& Prestige score & 200 & 1 & 0 \\
& Type school& 98 & 0 & 0 \\
\cline{2-5}
& Mean changes & 3.49 & 1.005 & \textbf{1} \\
\toprule
\multirow{10}{*}{Saheart} & Sbp & 200 & 8 & 37 \\
 & Tobacco & 200 & 132 & 94 \\
 & Ldl & 200 & 138 & 103 \\
 & Adiposity & 200 & 1 & 34 \\
 & Famhist & 129 & 83 & 59 \\
 & Typea & 200 & 12 & 33 \\
 & Obesity & 200 & 80 & 24 \\
 & Alcohol & 200 & 2 & 29 \\
 & Age & 0 & 103 & 0 \\
\cline{2-5}
 & Mean changes & 7.645 & 2.795 & \textbf{2.065} \\
\toprule
\multirow{9}{*}{Titanic} & Gender & 0 & 135 & 8 \\
& Age & 0 & 82 & 5 \\
& Class & 186 & 12 & 165 \\
& Embarked & 78 & 2 & 44 \\
& Country & 0 & 34 & 0 \\
& Fare & 200 & 19 & 25 \\
& Sibsp & 199 & 19 & 32 \\
& Parch & 200 & 8 & 32 \\
\cline{2-5}
& Mean changes & 4.315 & \textbf{1.555} & \textbf{1.555} \\
\toprule
\multirow{12}{*}{Wine} & Free acidity & 200 & 34 & 47 \\
& Volatile acidity & 200 & 108 & 64 \\
& Citric acid & 200 & 38 & 26 \\
& Residual sugar & 200 & 64 & 30 \\
& Chlorides & 200 & 43 & 16 \\
& Free sulphur dioxide & 200 & 93 & 59 \\
& Total sulphur dioxide & 200 & 24 & 40 \\
& Density & 200 & 52 & 15 \\
& PH & 200 & 10 & 34 \\
& Sulphates & 200 & 8 & 35 \\
& Alcohol & 200 & 77 & 89 \\
\cline{2-5}
& Mean changes & 11 & 2.755 & \textbf{2.275} \\
\toprule
\end{tabular}}
\label{tab_all_qual}
\end{center}
\end{table*}

Regarding Saheart, in the second part of Table \ref{tab_all_qual}, Growing spheres proposes many changes in all non-immutable features, which does not allow for any auditability. As for ONB-MACF and NICE, they give a reasonable amount of changes per instance, and both methods agree on the importance of the cumulative tobacco consumption and the amounts of low-density lipoprotein cholesterol, and to some extent the family history of such heart diseases. These changes make sense. NICE also gives much importance to Age (which is one of the immutable attributes) and Obesity. The model seems to pass the auditability test here too. No pairs of features are too common (ldl and tobacco being the most common with 13/200), and single changes in ldl and tobacco are the most common in both NICE and ONB-MACF (up to 31/200) but are still unusual.

As for Titanic, in the third part of Table \ref{tab_all_qual}, we again see that Growing spheres proposes changes in most non-immutable attributes most of the time. This time, ONB-MACF and NICE differ in the most relevant features, since ONB-MACF selects Class, while NICE points to Gender and Age. This is a relevant situation, since both Gender and Age are immutable features (and, in this case, bias-inducing) that the model must not use; thus, this model cannot pass the auditability test. This is so important that even the ONB-MACF method, which includes immutability clauses, could not find any prototypes with feasible changes that did not include changes in those features for some samples. In the case of Class, as selected by ONB, it makes sense that it would be an important feature while maintaining Gender and Age (as bribing or preference might have saved wealthy people). The most common pairs were age and gender on NICE (23/200) and class and embarked (13/200) and class and sibsp (12/200) on ONB-MACF, which were not too common. Single changes in gender on NICE and class on ONB-MACF were very common (up to 105/200).

Finally, the last part of Table \ref{tab_all_qual} shows the proposed changes on the Wine dataset. While Growing Spheres shows its inability to help regarding auditability, the feature changes are mainly mixed for ONB-MACF and NICE. Both prefer Volatile acidity, Free Sulphur Dioxide and Alcohol contents, but it is not as clear as in the other datasets. In this situation, checking the plausibility of the obtained counterfactuals would be useful. Counterfactuals obtained using the ONB-MACF method are plausible (as evidenced by the YNN metric). However, counterfactuals obtained using NICE are often too close to the boundary. 83/200 of the counterfactuals were borderline (2 or 3 of their neighbours were from the original instance's class), and 49/200 had 4 or 5 of their nearest neighbours with the original instance's class (which could be more comparable to adversarial examples). On the remaining 68 counterfactuals, the same three features remain the most important and in similar amounts, with a slight increase in the importance of Volatile acidity, so it is mostly a matter of distance rather than of overall counterfactual direction. This means that, in this dataset, counterfactuals follow a more case-by-case situation, without general rules for feature changes. There are no common pairs of features (the most common was alcohol + volatile acidity on NICE, which happened 8/200 times). Single changes in alcohol, volatile acidity or free sulphur dioxide were slightly common (up to 43/200) depending on the method.

\section{Concluding remarks} \label{concl}

This paper has presented ONB-MACF, a novel model-agnostic method to provide counterfactual explanations that simultaneously grasps data space coverage and class boundaries using data morphology. This method is designed to obtain close counterfactuals with few feature changes in a way that is adaptable to the characteristics of the dataset (feature restrictions, number of classes, etc.).

As exposed by the qualitative and quantitative analyses of the obtained benchmarking results, we have verified the usefulness of data-morphology-based strategies to provide counterfactual explanations. 
The ONB-MACF method has shown excellent results on multiple fronts simultaneously, such as closeness, sparsity, redundancy, actionability and validity, while also providing sensible counterfactuals that the audience of the model can understand and reason about. This was possible due to a solid model class boundaries estimation based on data morphology properties and the ONB-MACF method's careful design that allows it to foster explainability while complying with feature immutability and existing data distributions.

Finally, regarding future work lines, the method could be 
applied to interpretable models and, in particular, to rule-based systems, in order to improve the configuration and granularity of critical features and, with that, the efficacy of the global model. The method could also be adapted, combining it with deep metric learning or topological data analysis, to allow for its use on image datasets.
Other lines of future work could also include applications towards the auditability of models for ethics and fairness evaluation.

\section*{Acknowledgments}

This work has received funding from the Spanish Ministry of Science and Technology under project PID2020-119478GB-I00, including European Regional Development Funds, and the ``Cátedra Tecnalia en Inteligencia Artificial'' programme. It is also partially supported by the I+D+i project granted by C-ING-250-UGR23 co-funded by ``Consejería de Universidad, Investigación e Innovación'' and the European Union related to FEDER Andalucía Program 2021-27. J. Del Ser would like to thank the Basque Government for the funding support received through the EMAITEK and ELKARTEK programs (ref. KK-2023/00012), as well as the Consolidated Research Group MATHMODE (IT1456-22) granted by the Department of Education of this institution.

\bibliography{bibliography}
\bibliographystyle{spmpsci}

\vspace{12pt}

\appendix
\begin{section}{Experimental result tables for each dataset}\label{result_tables}

The benchmarking results (per dataset) of the different counterfactual methods in CARLA and ONB-MACF are presented in Tables \ref{tab_adult} to \ref{tab_wine}. In a few cases, no counterfactual was able to be generated for any of the 200 instances with a particular method, so the metrics could not be calculated - in these cases, their values are ``Na''. The mean of the scaled time results is presented in Table \ref{avg_time_tab}.

\begin{table*}[htbp]
\caption{Performance results of the studied counterfactual methods on the ``adult'' dataset, according to the benchmarking metrics}
\begin{center}
\scalebox{0.7}{
\begin{tabular}{cccccccccc}
\toprule
\textbf{Method}&\textbf{L0 norm}&\textbf{L1 norm}&\textbf{L2 norm}&\textbf{L-$\infty$ norm}&\textbf{Const. Vio.}&\textbf{Redun.}&\textbf{YNN}&\textbf{Succ. Rate} \\
\toprule
CEM & 5.785 & 0.925 & 0.840 & 0.831 & 0.645 & 3.960 & 0.311 & \textbf{1.000} \\

CEM-VAE & 5.785 & 0.925 & 0.840 & 0.831 & 0.645 & 3.960 & 0.312 & \textbf{1.000} \\
CLUE & 8.710 & 4.164 & 3.321 & 0.985 & 1.320 & 8.785 & 0.503 & \textbf{1.000} \\
CRUDS & 9.303 & 3.795 & 3.395 & 0.974 & 1.333 & 8.091 & 0.206 & 0.165 \\
DICE & 2.325 & 1.694 & 1.461 & 0.937 & 0.170 & \textbf{0.490} & \textbf{0.656} & \textbf{1.000} \\
FACE-EPSILON & 7.870 & 4.135 & 3.623 & 0.992 & 1.469 & 4.792 & 0.563 & 0.960 \\
FACE-KNN & 7.995 & 4.265 & 3.762 & 0.997 & 1.420 & 4.955 & 0.534 & \textbf{1.000} \\
GS & 6.300 & 1.513 & 1.315 & 0.859 & 0.100 & 3.920 & 0.330 & \textbf{1.000} \\
WACHTER & 6.660 & 0.864 & 0.669 & 0.668 & 0.840 & 4.415 & 0.211 & 0.530 \\
NICE & \textbf{1.365} & \textbf{0.217} & \textbf{0.058} & \textbf{0.175} & 0.130 & \textbf{0.025} & 0.314 & \textbf{1.000} \\
\hline
ONB-MACF & 2.765 & 0.274 & 0.091 & 0.203 & \textbf{0.000} & 1.425 & 0.544 & \textbf{1.000} \\
\hline

\end{tabular}}
\label{tab_adult}
\end{center}
\end{table*}

\begin{table*}[htbp]
\caption{Performance results of the studied counterfactual methods on the ``COMPAS'' dataset, according to the benchmarking metrics}
\begin{center}
\scalebox{0.7}{
\begin{tabular}{cccccccccc}
\toprule
\textbf{Method}&\textbf{L0 norm}&\textbf{L1 norm}&\textbf{L2 norm}&\textbf{L-$\infty$ norm}&\textbf{Const. Vio.}&\textbf{Redun.}&\textbf{YNN}&\textbf{Succ. Rate} \\
\toprule
CEM & 3.975 & 1.355 & 1.068 & 0.843 & 0.430 & 2.285 & 0.984 & \textbf{1.000} \\

CEM-VAE  &  3.970 & 1.336 & 1.040 & 0.843 & 0.475 & 2.275 & 0.984 & \textbf{1.000} \\
CLUE & 4.820 & 1.984 & 1.453 & 0.891 & 1.050 & 3.740 & 0.979 & \textbf{1.000} \\
CRUDS & 5.045 & 2.057 & 1.512 & 0.903 & 1.175 & 4.155 & \textbf{1.000} & \textbf{1.000} \\
DICE & 2.490 & 1.805 & 1.590 & 0.937 & 1.015 & 0.885 & 0.908 & \textbf{1.000} \\
FACE-EPSILON & 5.185 & 2.860 & 2.528 & 0.940 & 1.620 & 3.895 & 0.932 & \textbf{1.000} \\
FACE-KNN & 5.090 & 2.787 & 2.451 & 0.950 & 1.585 & 3.745 & 0.928 & \textbf{1.000} \\
GS & 3.760 & 0.961 & 0.794 & 0.763 & \textbf{0.025} & 0.935 & 0.448 & \textbf{1.000} \\
WACHTER & 4.669 & 7.957 & 130.542 & 2.653 & 0.790 & 1.828 & 0.575 & 0.785 \\
NICE & 2.395 & \textbf{0.896} & \textbf{0.681} & 0.648 & 1.105 & \textbf{0.340} & 0.155 & \textbf{1.000} \\
\hline
ONB-MACF & \textbf{2.245} & 1.060 & 0.853 & \textbf{0.642} & 0.095 & 0.620 & 0.201 & \textbf{1.000} \\
\hline

\end{tabular}}
\label{tab_COMPAS}
\end{center}
\end{table*}

\begin{table*}[htbp]
\caption{Performance results of the studied counterfactual methods on the ``Give Me Some Credit'' dataset, according to the benchmarking metrics}
\begin{center}
\scalebox{0.7}{
\begin{tabular}{cccccccccc}
\toprule
\textbf{Method}&\textbf{L0 norm}&\textbf{L1 norm}&\textbf{L2 norm}&\textbf{L-$\infty$ norm}&\textbf{Const. Vio.}&\textbf{Redun.}&\textbf{YNN}&\textbf{Succ. Rate} \\
\toprule
CEM & 6.950 & 0.956 & 0.724 & 0.688 & 0.670 & 5.990 & 0.945 & \textbf{1.000} \\

CEM-VAE & 6.950 & 0.958 & 0.725 & 0.689 & 0.670 & 5.990 & 0.946 & \textbf{1.000} \\
CLUE & 9.952 & 2.411 & 1.204 & 0.787 & 0.977 & 9.047 & 0.849 & 0.860 \\
CRUDS & 9.940 & 2.156 & 1.006 & 0.729 & 0.995 & 9.780 & \textbf{1.000} & \textbf{1.000} \\
DICE & \textbf{1.550} & 1.364 & 1.088 & 0.824 & 0.245 & \textbf{0.610} & 0.898 & \textbf{1.000} \\
FACE-EPSILON & 8.545 & 2.284 & 1.267 & 0.804 & 0.990 & 8.045 & 0.985 & \textbf{1.000} \\
FACE-KNN & 8.595 & 2.253 & 1.218 & 0.796 & 0.985 & 8.115 & 0.988 & \textbf{1.000} \\
GS & 8.965 & 0.748 & 0.577 & 0.584 & 0.155 & 6.490 & 0.522 & \textbf{1.000} \\
WACHTER & 9.735 & 1.032 & 0.594 & 0.563 & 0.750 & 6.785 & 0.691 & \textbf{1.000} \\
NICE & 3.410 & 0.497 & \textbf{0.137} & \textbf{0.311} & 0.090 & 1.150 & 0.417 & \textbf{1.000} \\
\hline
ONB-MACF & 2.255 & \textbf{0.459} & 0.147 & 0.337 & \textbf{0.005} & 0.795 & \textbf{1.000} & \textbf{1.000} \\
\hline

\end{tabular}}
\label{tab_give}
\end{center}
\end{table*}

\begin{table*}[htbp]
\caption{Performance results of the studied counterfactual methods on the ``HELOC'' dataset, according to the benchmarking metrics}
\begin{center}
\scalebox{0.7}{
\begin{tabular}{cccccccccc}
\toprule
\textbf{Method}&\textbf{L0 norm}&\textbf{L1 norm}&\textbf{L2 norm}&\textbf{L-$\infty$ norm}&\textbf{Const. Vio.}&\textbf{Redun.}&\textbf{YNN}&\textbf{Succ. Rate} \\
\toprule
CEM & 16.235 & 1.414 & 0.959 & 0.832 & \textbf{0.000} & 14.300 & 0.408 & \textbf{1.000} \\

CEM-VAE & 16.195 & 1.410 & 0.956 & 0.833 & \textbf{0.000} & 14.250 & 0.423 & \textbf{1.000} \\
CLUE & 21.621 & 7.563 & 4.577 & 1.058 & \textbf{0.000} & 20.514 & 0.678 & 0.180 \\
CRUDS & 21.025 & 7.503 & 6.516 & 1.459 & \textbf{0.000} & 20.740 & \textbf{1.000} & \textbf{1.000} \\
DICE & \textbf{1.805} & 1.731 & 1.406 & 0.921 & \textbf{0.000} & \textbf{0.335} & 0.515 & \textbf{1.000} \\
FACE-EPSILON & 18.075 & 3.606 & 1.865 & 0.934 & \textbf{0.000} & 15.400 & 0.825 & \textbf{1.000} \\
FACE-KNN & 18.230 & 3.485 & 1.800 & 0.939 & \textbf{0.000} & 15.580 & 0.809 & \textbf{1.000} \\
GS & 21.025 & 1.684 & 0.858 & 0.797 & \textbf{0.000} & 15.080 & 0.292 & \textbf{1.000} \\
WACHTER & 21.025 & 1.752 & 0.835 & 0.784 & \textbf{0.000} & 15.980 & 0.252 & \textbf{1.000} & \\
NICE & 5.555 & \textbf{0.537} & \textbf{0.095} & \textbf{0.201} & \textbf{0.000} & 1.435 & 0.483 & \textbf{1.000} \\
\hline
ONB-MACF & 3.62 & 0.773 & 0.248 & 0.314 & \textbf{0.000} & 0.655 & \textbf{1.000} & \textbf{1.000} \\
\toprule

\end{tabular}}
\label{tab_heloc}
\end{center}
\end{table*}

\begin{table*}[htbp]
\caption{Performance results of the studied counterfactual methods on the ``Irish'' dataset, according to the benchmarking metrics}
\begin{center}
\scalebox{0.7}{
\begin{tabular}{cccccccccc}
\toprule
\textbf{Method}&\textbf{L0 norm}&\textbf{L1 norm}&\textbf{L2 norm}&\textbf{L-$\infty$ norm}&\textbf{Const. Vio.}&\textbf{Redun.}&\textbf{YNN}&\textbf{Succ. Rate} \\
\toprule
CEM & 2.705 & 1.045 & 1.045 & \textbf{1.000} & 0.020 & 3.065 & 0.996 & \textbf{1.000} \\

CEM-VAE & 2.740 & 1.065 & 1.065 & \textbf{1.000} & 0.030 & 3.045 & 0.994 & \textbf{1.000} \\
CLUE & 5.000 & 3.001 & 2.679 & 1.069 & 1.000 & 11.905 & \textbf{1.000} & \textbf{1.000} \\
CRUDS & 4.828 & 3.729 & 3.895 & 1.428 & 1.000 & 9.630 & \textbf{1.000} & 0.960 \\
DICE & 1.680 & 1.410 & 1.301 & 1.005 & 0.075 & 0.370 & \textbf{1.000} & \textbf{1.000} \\
FACE-EPSILON & 4.092 & 2.509 & 2.062 & \textbf{1.000} & 0.563 & 4.134 & \textbf{1.000} & 0.710 \\
FACE-KNN & 4.010 & 2.418 & 1.969 & \textbf{1.000} & 0.550 & 3.885 & \textbf{1.000} & \textbf{1.000} \\
GS & 3.490 & 1.383 & 1.219 & \textbf{1.000} & \textbf{0.000} & 2.640 & \textbf{1.000} & \textbf{1.000} \\
WACHTER & 4.225 & 1.927 & 1.431 & 1.000 & 0.900 & 2.805 & \textbf{1.000} & \textbf{1.000} \\
NICE & 1.005 & 1.000 & 1.000 & \textbf{1.000} & \textbf{0.000} & \textbf{0.005} & \textbf{1.000} & \textbf{1.000} \\
\hline
ONB-MACF & \textbf{1.000} & \textbf{1.000} & \textbf{1.000} & \textbf{1.000} & \textbf{0.000} & 0.225 & \textbf{1.000} & \textbf{1.000} \\
\toprule

\end{tabular}}
\label{tab_irish}
\end{center}
\end{table*}

\begin{table*}[htbp]
\caption{Performance results of the studied counterfactual methods on the ``Saheart'' dataset, according to the benchmarking metrics}
\begin{center}
\scalebox{0.7}{
\begin{tabular}{cccccccccc}
\toprule
\textbf{Method}&\textbf{L0 norm}&\textbf{L1 norm}&\textbf{L2 norm}&\textbf{L-$\infty$ norm}&\textbf{Const. Vio.}&\textbf{Redun.}&\textbf{YNN}&\textbf{Succ. Rate} \\
\toprule
CEM & 7.770 & 0.859 & 0.329 & \textbf{0.414} & 0.975 & 5.965 & 0.195 & \textbf{1.000} \\

CEM-VAE & 7.770 & 0.855 & \textbf{0.328} & 0.417 & 0.970 & 5.895 & 0.201 & \textbf{1.000} \\
CLUE & Na & Na & Na & Na & Na & Na & Na & 0.000 \\
CRUDS & Na & Na & Na & Na & Na & Na & Na & 0.000 \\
DICE & \textbf{1.835} & 1.073 & 0.791 & 0.762 & 0.185 & \textbf{0.745} & 0.425 & \textbf{1.000} \\
FACE-EPSILON & 8.29 & 2.147 & 1.099 & 0.767 & 0.995 & 6.245 & 0.394 & \textbf{1.000} \\
FACE-KNN & 8.288 & 2.143 & 1.096 & 0.755 & 0.985 & 6.226 & 0.375 & 0.990 \\
GS & 7.645 & 1.053 & 0.704 & 0.687 & \textbf{0.000} & 4.235 & 0.394 & \textbf{1.000} \\
WACHTER & 9.000 & 29.22 & 158.312 & 4.994 & 1.000 & 1.743 & 0.000 & 0.565 \\
NICE & 2.795 & 0.943 & 0.652 & 0.624 & 0.515 & 0.905 & 0.623 & \textbf{1.000} \\
\hline
ONB-MACF & 2.065 & \textbf{0.838} & 0.520 & 0.565 & 0.005 & 0.815 & \textbf{1.000} & \textbf{1.000} \\
\toprule

\end{tabular}}
\label{tab_saheart}
\end{center}
\end{table*}

\begin{table*}[htbp]
\caption{Performance results of the studied counterfactual methods on the ``Titanic'' dataset, according to the benchmarking metrics}
\begin{center}
\scalebox{0.7}{
\begin{tabular}{cccccccccc}
\toprule
\textbf{Method}&\textbf{L0 norm}&\textbf{L1 norm}&\textbf{L2 norm}&\textbf{L-$\infty$ norm}&\textbf{Const. Vio.}&\textbf{Redun.}&\textbf{YNN}&\textbf{Succ. Rate} \\
\toprule
CEM & 3.745 & 1.190 & \textbf{0.902} & \textbf{0.747} & 1.805 & 1.955 & 0.202 & \textbf{1.000} \\

CEM-VAE & 3.760 & 1.210 & 0.921 & 0.750 & 1.825 & 1.965 & 0.199 & \textbf{1.000} \\
CLUE & 8.000 & 4.602 & 3.877 & 1.054 & 3.000 & 22.267 & 0.124 & \textbf{0.260} \\
CRUDS & Na & Na & Na & Na & Na & Na & Na & 0.000 \\
DICE & 1.830 & 1.688 & 1.648 & 0.966 & 0.950 & 0.580 & 0.471 & \textbf{1.000} \\
FACE-EPSILON & 5.556 & 3.040 & 2.729 & 0.995 & 2.556 & 5.297 & 0.874 & 0.900 \\
FACE-KNN & 5.470 & 3.039 & 2.738 & 0.993 & 2.525 & 5.255 & 0.890 & \textbf{1.000} \\
GS & 4.320 & 1.410 & 1.326 & 0.996 & \textbf{0.000} & 3.205 & 0.312 & \textbf{1.000} \\
WACHTER &7.430 & 3.451 & 2.948 & 1.000 & 2.690 & 8.595 & 0.194 & \textbf{1.000} \\
NICE & \textbf{1.555} & \textbf{1.048} & 0.983 & 0.879 & 1.255 & \textbf{0.160} & 0.775 & \textbf{1.000} \\
\hline
ONB-MACF & \textbf{1.555} & 1.205 & 1.128 & 0.962 & 0.065 & 1.225 & \textbf{0.960} & \textbf{1.000} \\
\toprule

\end{tabular}}
\label{tab_titanic}
\end{center}
\end{table*}

\begin{table*}[htbp]
\caption{Performance results of the studied counterfactual methods on the ``Wine'' dataset, according to the benchmarking metrics}
\begin{center}
\scalebox{0.7}{
\begin{tabular}{cccccccccc}
\toprule
\textbf{Method}&\textbf{L0 norm}&\textbf{L1 norm}&\textbf{L2 norm}&\textbf{L-$\infty$ norm}&\textbf{Const. Vio.}&\textbf{Redun.}&\textbf{YNN}&\textbf{Succ. Rate} \\
\toprule
CEM & 10.955 & 0.259 & 0.067 & 0.185 & \textbf{0.000} & 9.875 & 0.610 & \textbf{1.000} \\

CEM-VAE & 10.955 & 0.253 & 0.064 & 0.184 & \textbf{0.000} & 9.900 & 0.613 & \textbf{1.000} \\
CLUE & 11.000 & 3.828 & 2.726 & 1.172 & \textbf{0.000} & 9.745 & 0.588 & \textbf{1.000} \\
CRUDS & 11.000 & 1.574 & 0.467 & 0.499 & \textbf{0.000} & 10.155 & \textbf{1.000} & \textbf{1.000} \\
DICE & \textbf{1.565} & 0.676 & 0.386 & 0.525 & \textbf{0.000} & 0.575 & 0.833 & \textbf{1.000} \\
FACE-EPSILON & 10.785 & 1.065 & 0.193 & 0.277 & \textbf{0.000} & 8.600 & 0.891 & \textbf{1.000} \\
FACE-KNN & 10.800 & 1.093 & 0.211 & 0.288 & \textbf{0.000} & 8.530 & 0.866 & \textbf{1.000} \\
GS & 11.000 & 0.242 & \textbf{0.014} & 0.055 & \textbf{0.000} & 7.500 & 0.422 & \textbf{1.000} \\
WACHTER & 11.000 & 0.313 & 0.014 & \textbf{0.029} & \textbf{0.000} & 8.645 & 0.560 & \textbf{1.000} \\
NICE & 2.755 & \textbf{0.177} & 0.023 & 0.111 & \textbf{0.000} & 0.620 & 0.524 & \textbf{1.000} \\
\hline
ONB-MACF & 2.275 & 0.242 & 0.039 & 0.142 & \textbf{0.000} & \textbf{0.490} & \textbf{1.000} & \textbf{1.000} \\
\toprule

\end{tabular}}
\label{tab_wine}
\end{center}
\end{table*}

\begin{table*}[htbp]
\caption{Average of the time results over the 8 datasets, where times for each dataset were transformed so that the scale went from 0 to 1 and
so higher values indicated better performance.}
\begin{center}
\scalebox{0.7}{
\begin{tabular}{cccccccccc}
\toprule
\textbf{Method}&\textbf{Average Time} \\
\toprule
CEM & 0.880 \\
CEM-VAE & 0.876 \\
CLUE & 0.696 \\
CRUDS & 0.328 \\
DICE & 0.982 \\
FACE-EPSILON & 0.894 \\
FACE-KNN & 0.899 \\
GS & 0.998 \\
WACHTER & 0.645 \\
NICE & 0.999 \\
\hline
ONB-MACF & 0.504 \\
\toprule

\end{tabular}}
\label{avg_time_tab}
\end{center}
\end{table*}

\end{section}

\end{document}